\documentclass{ieeeaccess}
\usepackage{cite}
\usepackage{amsmath,amssymb,amsfonts}
\usepackage{algorithmic}
\usepackage{graphicx}
\usepackage{textcomp}
\usepackage{cite}
\usepackage{booktabs}
\usepackage{multirow}
\usepackage{graphicx}
\usepackage{bm}
\usepackage{booktabs}
\usepackage{subfigure}
\usepackage{xcolor}
\usepackage{hyperref}
\usepackage[capitalize]{cleveref}
\usepackage{array}      
\usepackage{soul}
\soulregister{\cite}{1}
\soulregister{\cref}{1}
\soulregister{\ref}{1}


\makeatletter
\AtBeginDocument{\DeclareMathVersion{bold}
\SetSymbolFont{operators}{bold}{T1}{times}{b}{n}
\SetSymbolFont{NewLetters}{bold}{T1}{times}{b}{it}
\SetMathAlphabet{\mathrm}{bold}{T1}{times}{b}{n}
\SetMathAlphabet{\mathit}{bold}{T1}{times}{b}{it}
\SetMathAlphabet{\mathbf}{bold}{T1}{times}{b}{n}
\SetMathAlphabet{\mathtt}{bold}{OT1}{pcr}{b}{n}
\SetSymbolFont{symbols}{bold}{OMS}{cmsy}{b}{n}
\renewcommand\boldmath{\@nomath\boldmath\mathversion{bold}}}
\makeatother

\renewcommand{\thetable}{\Roman{table}}

\crefname{table}{Table}{Tables}
\Crefname{table}{Table}{Tables}
\crefname{figure}{Fig.}{Fig.}
\Crefname{figure}{Fig.}{Fig.}

\makeatletter
\newcommand{\DDUNet}[3]{\ensuremath{\text{DDU-Net}(#1, #2, \text{#3})}}
\newcommand{\UNet}[1]{\ensuremath{\text{U-Net}(#1)}}

\def\BibTeX{{\rm B\kern-.05em{\sc i\kern-.025em b}\kern-.08em
    T\kern-.1667em\lower.7ex\hbox{E}\kern-.125emX}}

\begin{document}
\history{Received March 11, 2025, Accepted March 30, 2025. Date of publication April 15, 2025, date of current version April 22, 2025.}
\doi{10.1109/ACCESS.2025.3561033}

\title{DDU-Net: A Domain Decomposition-Based CNN for High-Resolution Image Segmentation on Multiple GPUs}
\author{\uppercase{Corné Verburg}\authorrefmark{1},
\uppercase{Alexander Heinlein}\authorrefmark{1}, and \uppercase{Eric. C. Cyr}\authorrefmark{2},}

\address[1]{Delft Institute of Applied Mathematics, Delft University of Technology, Delft, 2628 CD, Zuid-Holland, The Netherlands}
\address[2]{Computational Mathematics Department, Sandia National Laboratories, Albuquerque, New Mexico, USA}

\markboth
{C. Verburg \headeretal: DDU-Net: A Domain Decomposition-Based CNN for High-Resolution Image Segmentation}
{C. Verburg \headeretal: DDU-Net: A Domain Decomposition-Based CNN for High-Resolution Image Segmentation}

\corresp{Corresponding authors: A. Heinlein (a.heinlein@tudelft.nl) and C. Verburg (c.verburg@tudelft.nl)}

\begin{abstract}
The segmentation of ultra-high resolution images poses challenges such as loss of spatial information or computational inefficiency. In this work, a novel approach that combines encoder-decoder architectures with domain decomposition strategies to address these challenges is proposed. Specifically, a domain decomposition-based U-Net (DDU-Net) architecture is introduced, which partitions input images into non-overlapping patches that can be processed independently on separate devices. A communication network is added to facilitate inter-patch information exchange to enhance the understanding of spatial context. Experimental validation is performed on a synthetic dataset that is designed to measure the effectiveness of the communication network. Then, the performance is tested on the DeepGlobe land cover classification dataset as a real-world benchmark data set. The results demonstrate that the approach, which includes inter-patch communication for images divided into $16\times16$ non-overlapping subimages, achieves a $2-3\,\%$ higher intersection over union (IoU) score compared to the same network without inter-patch communication. The performance of the network which includes communication is equivalent to that of a baseline U-Net trained on the full image, showing that our model provides an effective solution for segmenting ultra-high-resolution images while preserving spatial context. The code is available at~\href{https://github.com/corne00/DDU-Net}{\texttt{https://github.com/corne00/DDU-Net}}.
\end{abstract}


\titlepgskip=-21pt

\maketitle

\section{Introduction}
\label{sec:introduction}
\PARstart{T}{he} vast majority of deep learning models in computer vision focuses on low-resolution 2D and 3D images, typically $256 \times 256$ pixels or smaller. However, the increased utilization of high-resolution image datasets introduces new challenges due to the memory constraints of a single GPU, especially for memory-intensive tasks such as semantic segmentation of images~\cite{bakhtiarnia2024efficient}. Semantic segmentation is the computer vision task of classifying the pixels in the input into distinct, non-overlapping semantic categories. Ultra-high-resolution image segmentation holds significance in diverse fields such as object segmentation in satellite images~\cite{korznikov2021using,li2024glh}, metallic surface defect detection~\cite{tao2018automatic,lv2023fast}, and computer-aided medical diagnosis~\cite{asgari2021deep,bazin2014computational}. Whereas deep convolutional neural networks (CNNs) have achieved remarkable success in image segmentation \cite{lecun2015deep}, most of these models are unsuitable for model training and inference for high-resolution images due to their high memory requirements.

To illustrate this, consider computed tomography (CT) scans with sub-millimeter resolution, resulting in voxel image data with a typical resolution of $512\times512\times512$ voxels. Even with half-precision floating-point numbers and a modest batch size of 8, processing such images with a 1-layer convolutional neural network with 64 filters demands over 137GB of GPU/TPU memory, as highlighted in \cite{hou2019high}. Dealing with such high-resolution inputs using conventional strategies, like down-sampling or patch cropping, generally leads to the loss of detailed information or spatial context, resulting in a lower segmentation accuracy~\cite{chen2019collaborative,tsaris2023scaling,hou2019high, bakhtiarnia2024efficient}. 

In this paper, we propose a novel network architecture based on the well-known U-Net~\cite{ronneberger2015u} developed for biomedical image segmentation tasks. The U-Net is a convolutional neural network (CNN) with a bottleneck structure and skip connections between the encoder and decoder paths. This foundational architecture and variations of it~\cite{cciccek20163d,zhou2018unet++,oktay2018attention,diakogiannis2020resunet} have demonstrated remarkable, state-of-the-art accuracy in semantic segmentation and other image-to-image tasks~\cite{siddique2021u}. However, its substantial memory requirements are prohibitive for application to high-resolution images on computational devices with limited memory~\cite{azad2022medical}. The high memory demands for training stem from the required storage of intermediate feature maps obtained during forward propagation to be used during the subsequent backward propagation pass. This is especially acute when dealing with high-resolution inputs, as it leads to high-dimensional feature maps in the first and last few layers of the network. 

Our proposed model integrates the established U-Net architecture with a divide-and-conquer strategy inspired by domain decomposition methods (DDMs)~\cite{toselli_domain_2005} to deal with computational device memory limitations. Previous parallelization strategies employ a decomposition of the image into subimages (subdomains)~\cite{hou2019high,seal2020toward,he2016deep}, with communicated margins before each convolution or redundant computations providing global contextual information between subimages.
Our approach similarly decomposes the image. However, communication is explicitly limited to only the bottleneck within the U-Net architecture. This minimizes the communication overhead while preserving essential contextual information.

We summarize our main contributions as follows:
\begin{itemize}
    \item We propose a novel approach combining U-Net architectures with domain decomposition strategies to segment ultra-high-resolution images efficiently while preserving spatial context. 
    \item We show that the communication network, which is an important component of our approach, can be employed to exchange information between different subimages. It enhances understanding spatial context without significant computational overhead and with minimal extra communication and memory cost. 
    \item By evaluating our architecture on a synthetic and a realistic image dataset, we demonstrate competitive segmentation performance compared to baseline U-Net models. Our approach remains scalable, even when training is confined to the largest image portion that the available devices can handle.
\end{itemize}

The remainder of this paper is organized as follows: In~\cref{sec:related-work}, we discuss the U-Net architecture in more detail as well as existing memory parallelization strategies. Furthermore, we highlight approaches using ideas from the field of domain decomposition to speed up and parallelize CNNs.  \Cref{sec:methodology} introduces our novel parallel architecture and training methodology as well as the considered test datasets. In~\cref{sec:architecture}, we discuss the influence of this new architecture on memory requirements and the receptive field. Then, we evaluate our proposed network model and the related training strategy using these datasets in~\cref{sec:experimental-results}. We provide experimental results for the different datasets in terms of the segmentation accuracy of our approach compared to global semantic segmentation U-Net models. Finally, in~\cref{sec:conclusion}, we conclude and present possible future research directions. 

\section{Related Work}
\label{sec:related-work}
In this section, we discuss the U-Net, which serves as the basis for our proposed approach, and the receptive field, a key concept for understanding CNNs. Additionally, we provide an overview of past efforts to tackle the memory challenges associated with the U-Net and other CNN-based semantic segmentation models; we discuss both strengths and drawbacks of these methods. Finally, we introduce the idea of domain decomposition as a natural way to handle memory constraints, highlighting some previous works that adopt this approach for both classification and  segmentation.

\subsection{The U-Net} \label{subsec:the-u-net}
The new segmentation approach introduced in the this paper is based on the U-Net architecture~\cite{ronneberger2015u}. Since its publication, many variants and extensions of the U-Net architecture have been developed, for instance, 3D U-Net~\cite{cciccek20163d}, UNet++~\cite{zhou2018unet++}, Attention U-Net~\cite{oktay2018attention}, and ResUNet-a~\cite{diakogiannis2020resunet}. We assume that the reader is familiar with the fundamental concepts underlying CNNs~\cite{lecun1989generalization} and the building blocks of such models -- convolutional and pooling layers. For a detailed explanation of these concepts, we refer to the rich literature on this topic, for instance,~\cite[Chapt.~9]{Goodfellow-et-al-2016} or~\cite[Chapt.~10]{bishop_deep_2024}. 

\begin{figure}[htbp]
  \centering
  \includegraphics[width=1.0 \linewidth]{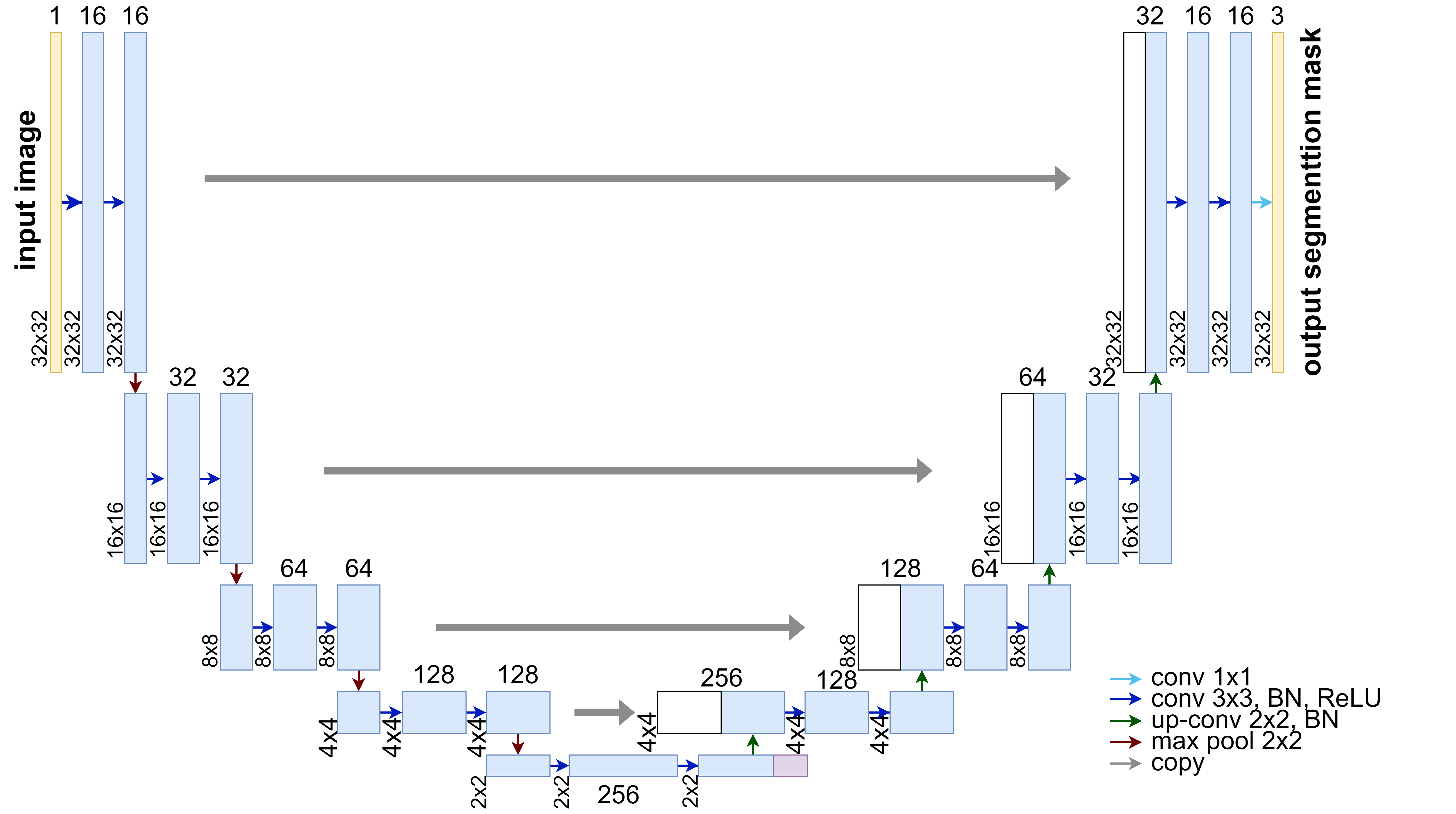}
  \caption{U-Net architecture for 32$\times$32 pixel input images and corresponding masks. Each blue block represents a multi-channel feature map, with resolutions indicated at the lower left edge of each box. White boxes show copied feature maps from the skip connections (gray arrows). The colored arrows denote different operations. This figure is based on the architecture described in~\cite{ronneberger2015u}. Image is best viewed online.}
   \label{fig:unet}
\end{figure}

The U-Net architecture \cite{ronneberger2015u}, depicted in~\cref{fig:unet}, consists of two pathways: the \textit{contraction path} or encoder path, and the \textit{expansive path} or decoder path. The contraction path involves repeated blocks, each consisting of two successive $3 \times 3$ convolutions, followed by a ReLU activation and a max-pooling layer. Conversely, the expansion path employs blocks that up-sample the feature map using $2 \times 2$ transposed convolutions. Subsequently, the feature maps from the corresponding layer in the contracting path are cropped (if the up-sampling path discards the boundary pixels) and concatenated to the up-sampled feature maps. This concatenation is followed by two successive $3 \times 3$ convolutions and a ReLU activation. 

In the final stage, the number of feature maps is reduced to the desired number of classes by a $1 \times 1$ convolution, producing the segmented image. The \textit{skip connections} between the contraction path, which captures (global) context and features, and the expansive path, which enables precise, fine-grained localization, distinguishes the U-Net from other related network architectures, such as the Fully Convolutional Network (FCN) presented in \cite{long2015fully}. Note that the originally proposed U-Net returns a cropped segmentation mask (due to the loss of border pixels in every convolution). However, more recent implementations of the U-Net often use zero padding at the borders to obtain output dimensions equal to the input dimensions, such that the input dimensions are preserved and cropping is not required; cf.~\cite{cathelain2020u, zhang2023semantic,bangaru2022scanning}.

\subsection{The Receptive Field}
An important concept for understanding CNNs is their \textit{receptive field}, or \textit{field of view}. The receptive field of a CNN is the area of the input image that influences the output of a pixel in the CNN output. Unlike fully-connected neural networks, where each input neuron is connected with each output neuron, CNNs have outputs that depend on specific regions of the input layer. In each layer of the CNN, the receptive field expands as we move deeper into the network, allowing higher-level features to be influenced by larger portions of the input image.

The theoretical size of the receptive field can be analyzed both experimentally and theoretically, as has been discussed extensively in~\cite{richter2022receptive}. The size of the receptive field significantly impacts the predictive performance of the CNN. If the receptive field is too small, the network may not capture all relevant information, leading to suboptimal predictions. For tasks like segmentation, the receptive field should be large enough to include all pixels in the input image that are relevant for predicting the correct pixel class. Note that the optimal size of the receptive field depends on both the network architecture as well as the data on which the network is trained.

In practice, the receptive field size of a CNN is influenced by several factors, including the number of convolutional layers, the size of the convolutional kernels, and the presence of pooling layers; cf.~\cite{luo2016understanding}. Increasing the depth of the network and the size of the kernels can expand the receptive field, allowing the network to capture more context from the input image. However, this also increases computational complexity and the risk of overfitting.

\subsection{Overcoming U-Net Memory Limits}
\label{subsec:Overcoming-U-Net-Memory-Limits}
One issue shared by most U-Net variants is the large memory requirements due to the storage of intermediate feature maps during training, making the model unsuitable for high-resolution applications with limited-memory computing devices~\cite{azad2022medical}. 

Two naive methods to limit memory usage during U-Net training are image patching and image downsizing. To show the downsides of these approaches, we trained a ResNet18-UNet \mbox{\cite{diakogiannis2020resunet}} on the DeepGlobe land cover classification dataset. For more details on the training procedure and dataset used, we refer to Section{~\ref{sec:methodology}}. Image patching involves training the U-Net model on randomly extracted patches from the full-resolution image. The patch size can be chosen based on the available memory and the optimal batch size. However, this approach does not allow the network to see the patch in a broader spatial context, which may be important. This is illustrated in Table{~\ref{tab:random-patching}}, where a ResNet18-U-Net  was trained on patches of varying sizes. As shown, larger patch sizes (which include more context) lead to significantly better predictions, both on the training and test datasets.

\begin{table}[ht]
\centering
\begin{tabular}{@{}ccc@{}}
\toprule
\textbf{Train patch size} & \textbf{mean IoU (train)} & \textbf{mean IoU (test)} \\ \midrule
16 & 0.3866 & 0.3099 \\
32 & 0.4651 & 0.3943 \\
64 & 0.5371 & 0.4882 \\
128 & 0.5943 & 0.5691 \\
256 & 0.6741 & 0.6342 \\
512 & 0.7383 & 0.6795 \\
1024 &\textbf{ 0.7601} & \textbf{0.6847} \\ \bottomrule
\end{tabular}%
\caption{Random patching results on the DeepGlobe dataset. All IoU scores are obtained on the full scale images ($2\,048\times 2\,048$ pixels). Training happened on different patch sizes. Training and methodology details are outlined in more detail in Section{~\ref{sec:methodology}}.}
\label{tab:random-patching}
\end{table}

A second naive approach to address memory issues is downsampling. By reducing the resolution of the images, memory usage during training can be decreased. However, this results in the loss of fine-grained details, as these are discarded in the downsampling process. While this method yields better results than using random patches of small sizes (see Table{~\ref{tab:downsampling}}), we still observe that prediction accuracy is highest when using the largest training images, thus preserving as much fine-grained context as possible.

\begin{table}[ht]
\centering
\begin{tabular}{ccc}
\hline
\textbf{Downsampled size} & \textbf{mean IoU (train)} & \textbf{mean IoU (test)} \\ \hline
16 & 0.5001 & 0.4654 \\
32 & 0.5875 & 0.5058 \\
64 & 0.6688 & 0.5601 \\
128 & 0.7194 & 0.6582 \\
256 & 0.7578 & 0.6896 \\
512 & \textbf{0.7740} & 0.6923 \\
1024 & 0.7619 & \textbf{0.6926} \\ \hline
\end{tabular}%
\caption{Downscaling results on the DeepGlobe dataset: All IoU scores were obtained on the full-scale images ($2048 \times 2048$ pixels), while training was conducted on patches of varying sizes. Training and methodology details are outlined in more detail in Section{~\ref{sec:methodology}}.}
\label{tab:downsampling}
\end{table}

A natural approach to overcome these constraints 
is to partition the memory load of the feature maps over different computational devices by using data and/or model parallelism. Several parallelized forms of the U-Net employ parallelism to reduce the memory per device. Both in~\cite{hou2019high} and in~\cite{tsaris2021distributed}, the authors implement a spatial partitioning technique that decomposes the input and output of the convolutional layers into smaller, non-overlapping subimages. Before each convolution operation, devices exchange patch margins of the feature maps of half the convolution kernel size with each other. While this approach introduces a communication overhead, it effectively distributes memory across devices without altering the fundamental architecture of the U-Net. Our approach, on the other hand, involves architectural modifications to the U-Net to further optimize performance and limit the communication to the bottleneck of the architecture.

The authors of~\cite{seal2020toward} partition the image into overlapping subimages, with the overlap size determined by the receptive field size. Due to redundant computations ($\mathcal{O}(q\frac{4\epsilon}{N})$ for an $N \times N$ pixel image partitioned into $q \times q$ subimages and a U-Net with receptive field size $\epsilon \times \epsilon$), this approach allows for a fully parallelized execution of both forward and backward passes. \cite{tsaris2023scaling} adopts a similar strategy, but with an application to image classification rather than image segmentation using the ResNet architecture~\cite{he2016deep}.

These approaches have been successful in partitioning the U-Net in such a way that the model can be trained and evaluated in parallel, even for ultra-high-resolution image datasets. However, they either involve communicating margins before each convolutional operation (this is the case for~\cite{hou2019high,tsaris2021distributed}), which leads to communication overhead through the many point-to-point messages, or they entail many redundant computations (this is the case for~\cite{he2016deep,tsaris2023scaling}). 

\subsection{Memory Optimization for CNNs Using Domain Decomposition Approaches} 
For improving memory efficiency, DDMs are inspiring due to their inherent parallelization and scalability properties, resulting from localization of the computations. DDMs are effective iterative solvers for (discretized) partial differential equations (PDEs), exhibiting scalability through a divide-and-conquer strategy that partitions the computational domain into overlapping or non-overlapping subdomains; see, for instance,~\cite{chan1994domain,herrera2003domain,toselli_domain_2005,dolean2015introduction}). The PDE problem is partitioned into subproblems defined on these subdomains. This enables parallel execution of computations within the subdomains. Whereas global convergence is ensured by well-balanced neighbor communication at potentially overlapping subdomain boundaries, and a limited amount of global communication. Highly scalable state-of-the-art DDMs are, for instance, variants of overlapping Schwarz methods~\cite{dohrmann2008domain,spillane_abstract_2014} or balancing domain decomposition by constraints (BDDC) \cite{cros_preconditioner_2003,dohrmann_preconditioner_2003} and finite element tearing and interconnecting - dual primal (FETI-DP) \cite{farhat_feti-dp_2001,farhat_scalable_2000}.

Recently, there has been an increasing interest in integrating DDMs with machine learning (ML) algorithms, combining the strengths of both fields~\cite{heinlein2021combining,klawonn2023machine}. Up to now, the majority of works focuses on combining DDMs and ML for solving PDEs. The present paper, however, pursues the combination of CNN with DDM techniques for semantic image segmentation. Notably, to the best of the authors' knowledge, no prior studies have tackled image segmentation tasks using ML explicitly based on domain decomposition strategies, although some methods have been proposed that have similarities with domain decomposition strategies. Existing research mainly focuses on image classification, which is closely related to semantic segmentation; however, due to the different model output, network architectures are somewhat different. Here, we provide a concise overview of methods combining DDM strategies with ML for image classification and semantic image segmentation tasks.

\begin{figure*}[ht!]
  \centering
  \includegraphics[width=1.0\linewidth]{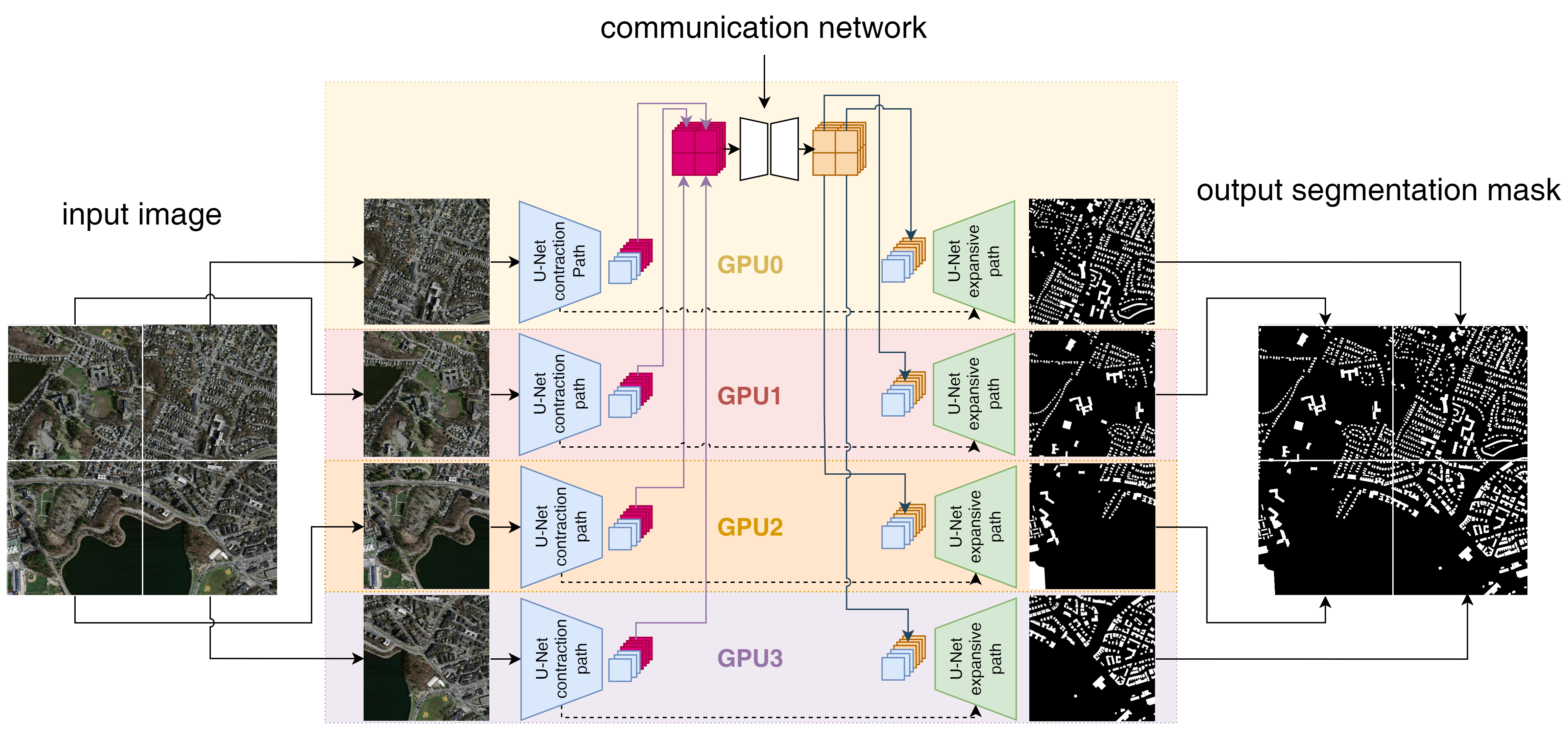}
  \caption{Schematic of the proposed network architecture. Input images are partitioned into subimages that are processed independently in the encoder paths. After encoding, a number of encoded feature maps is communicated to the device containing the communication network and then processed via the communication network. The output of this network replaces the input feature maps. The decoding is also done in parallel without communication between the computational devices. Dashed arrows indicate skip connections. Detailed architectures of the encoder-decoder network and communication network are shown in~\cref{fig:proposed_subnetwork,fig:proposed_communication_network}, respectively.}
  \label{fig:proposed-architecture}
\end{figure*}

In~\cite{man2023multi}, the authors propose a image partitioning approach for image segmentation, reducing network complexity and enhancing parallelizability. They partition input images into non-overlapping subimages and subsequently train smaller local CNNs on these subimages instead of a large global CNN. This enhances model specialization to specific image regions and reduces overall complexity without a significant impact on test errors compared to a global CNN. Importantly, the authors process the subimages completely independently, with no coupling between them.
In~\cite{klawonn2023domain}, DDM strategies are applied to train a CNN-DNN (convolutional neural network - deep neural network) architecture for image classification. The authors partition the input image into subimages, each used to train an individual CNN for predicting the class locally. A DNN then aggregates these local predictions into one global prediction. The authors interpret the DNN as a coarse problem solver that combines the finer-grained information from the local networks. In~\cite{gu2022decomposition}, the authors decompose a global CNN for image classification into a finite number of smaller, independent local subnetworks along its width. The weights obtained from training these local subnetworks serve as an initialization for the subsequent training of the global network, employing a transfer learning strategy. In \cite{mills2019extensive}, the authors propose a physically-motivated neural network topology for estimating extensive parameters such as energy or entropy. They decompose the domain into subdomains with \textit{focus} and \textit{context} regions, overlapping based on the \textit{locality} of the extensive parameter estimated. The same subnetwork weights are used on the different subdomains (we employ this strategy as well in our approach), leading to a reduction of the number of parameters, enhanced parallelizability, and faster inference. Finally, in \cite{park2019small}, a DDM-inspired segmentation algorithm is proposed that divides the input image into several overlapping subimages and trains one segmentation network on these subimages, allowing for parallel inference. The authors conclude that this approach leads to better segmentation accuracy of small objects.

\section{Methodology}
\label{sec:methodology}
In this section, we introduce our domain decomposition-based U-Net (DDU-Net) architecture, along with our training approach. We also describe the datasets used for testing the model and outline the model training procedure.

\subsection{Network architecture} \label{subsec:network_architecture}
\subsubsection{Decomposition of the images and masks}\label{subsub:decomposition-of-image-domain}
\Cref{fig:proposed-architecture} shows a schematic representation of our proposed network architecture. Starting with a dataset of (high-resolution) images and corresponding segmentation masks, the model processes a 2D pixel image with dimensions $H \times W$ (left) and outputs a probability distribution for $K \in \mathbb{N}$ classes for each pixel of the image (right). 

Following DDM strategies, we decompose the input data into $N \times M$ non-overlapping subimages, with $N, M \in \mathbb{N}$. In particular, each image is divided into $N \times M$ subimages with heights $H_{i}$ and widths $W_{j}$, $i=1,\ldots, N$ and $j=1,\ldots,M$, such that $\sum_{i=1}^{N} H_i = H$ and $\sum_{j=1}^{M} W_i = W$. The corresponding segmentation masks are partitioned similarly. The subimages and sub-masks are distributed across the computational devices (e.g., GPUs/TPUs) to balance the workload evenly.

After partitioning, the subimages are processed independently, in parallel, by the encoders on the separate computational devices. For more details on the architectures of encoders and decoders, we refer to~\cref{subsub:subnetwork}. A chosen number of feature maps from the last layer of the encoder is sent to the communication network (see~\cref{subsub:communication-network}), allowing for the exchange of relevant context between subimages. The communication network concatenates the encoded feature maps from all subimages corresponding to the arrangement of the subimages in the original full resolution image and processes them. Each output feature map of the communication network is again partitioned into $N \times M$ sub-feature maps, that are then sent back to the devices. The decoders then produce sub-predictions for each subimage, which are concatenated to form a global predicted mask for the entire high-resolution image.

\subsubsection{Encoder-decoder networks} \label{subsub:subnetwork}
\begin{figure}[hb!]
    \centering
    \includegraphics[width=1.0\linewidth]{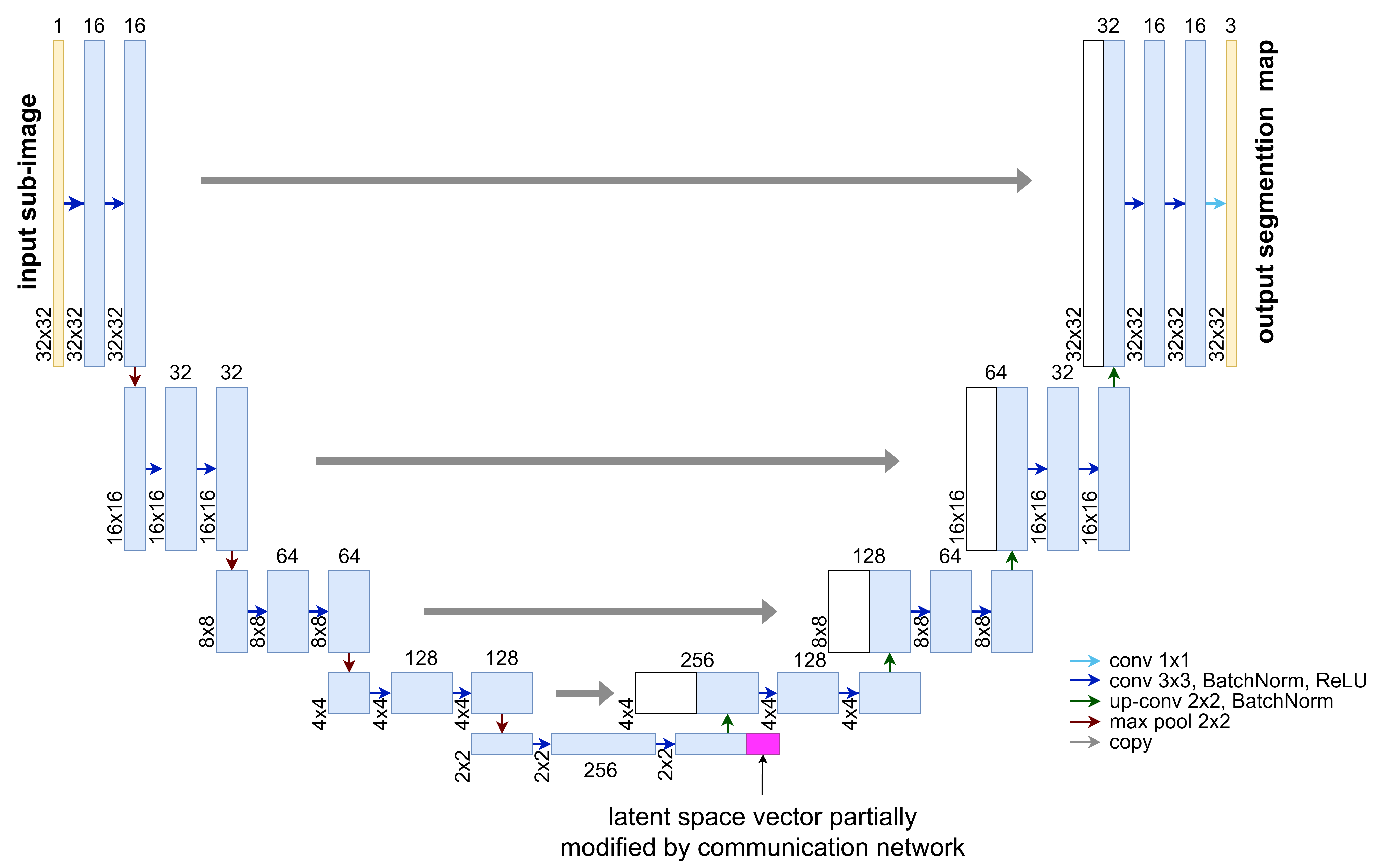}
    \caption{The proposed encoder-decoder architecture. The architecture of the encoder-decoder is nearly identical to the architecture of U-Net \cite{ronneberger2015u}. The only difference is located in the latent space vector, where a number of the feature maps are modified by the communication network. For optimal detail resolution, view this figure on a digital device.}
    \label{fig:proposed_subnetwork}
\end{figure}

\Cref{fig:proposed_subnetwork} shows the encoder-decoder architecture used in this work, which closely aligns with the classical U-Net architecture \cite{ronneberger2015u}; cf.~\cref{fig:unet}. The primary difference is a modification in the deepest layer of the encoder path to facilitate communication between the local encoder-decoder clones, which will be discussed below.

As discussed in~\cref{subsub:decomposition-of-image-domain}, each computational device contains a separate encoder-decoder network, allowing for parallel processing of the subimages. However, these networks share their weights to ensure consistent segmentation across subimages, making them local clones of a global encoder-decoder network. This implies that during training, the weights and gradients need to be synchronized after each backward propagation step.

Each local encoder network processes its corresponding input subimage with several blocks of $3 \times 3$ convolutions, each followed by a batch normalization layer and a ReLU activation. During training, a dropout layer is also applied to prevent over-fitting. After two convolution blocks, max-pooling is performed to reduce the spatial dimension of the feature maps. The deepest encoder layer produces $256$ spatially coarse feature maps (this number can vary with changes in the network architecture), some of which are employed for communication between the encoder-decoder networks. The output of each encoder network are $256$ feature maps, with small spatial dimensions. The last $F$ of these 256 feature maps are directed to the communication network, which processes them and returns modified feature maps with the same dimensions that replace the original input feature maps; see~\cref{subsub:communication-network}. After communicating, the feature maps then progress through the expansive path, resulting in a final segmented sub-mask for each subimage. These sub-masks are concatenated to form a full mask covering the entire full-resolution image.

\subsubsection{Communication network}\label{subsub:communication-network}
\begin{figure}[ht!]
    \centering
    \includegraphics[width=1.0\linewidth]{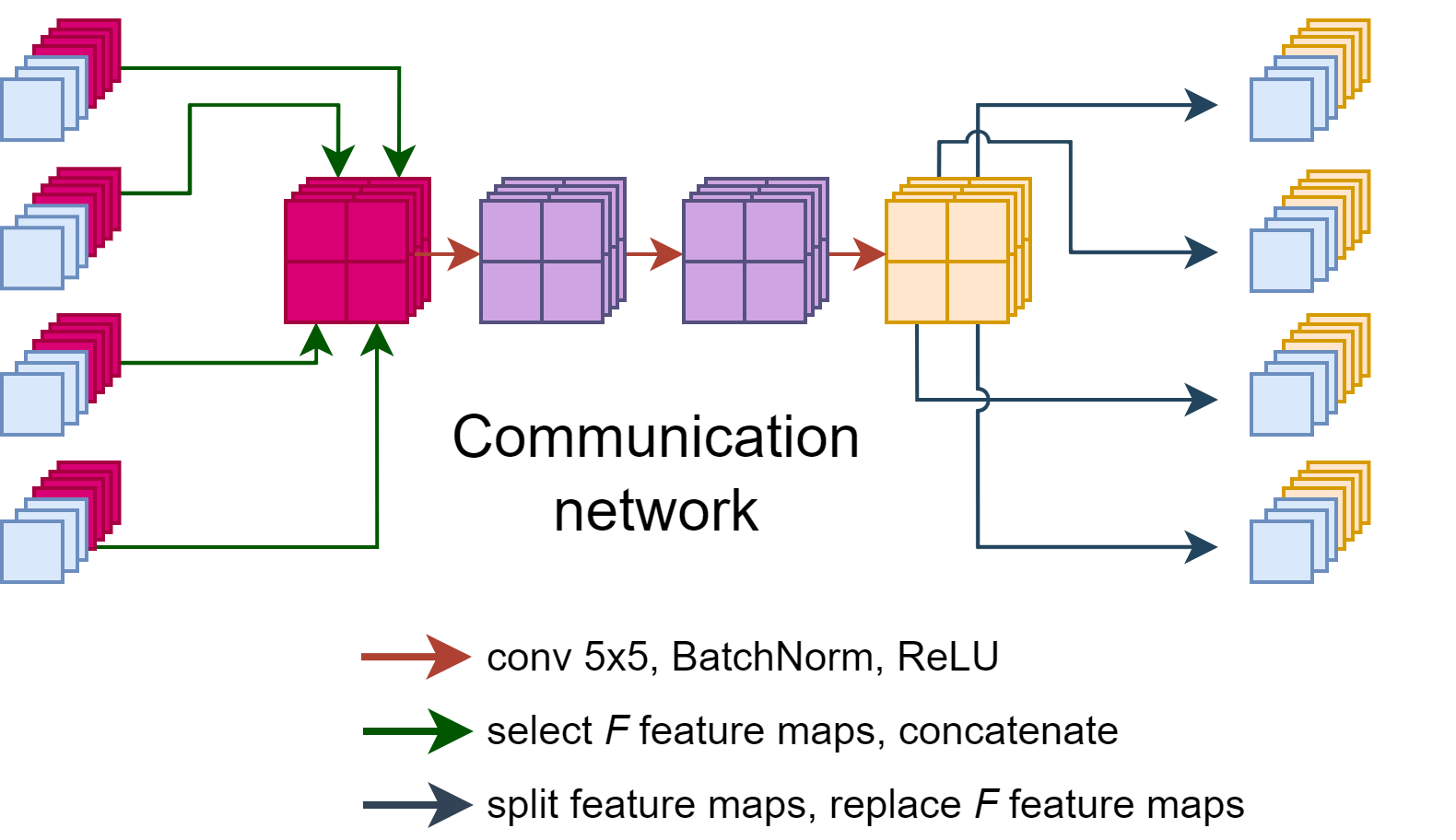}
    \caption{The proposed communication network for four subimages.} 
    \label{fig:proposed_communication_network}
\end{figure}

The communication network, depicted in~\cref{fig:proposed_communication_network}, transfers contextual features, captured by the encoders operating on subimages, between the encoded subimage feature maps. 

For this paper, the communication network consists of three layers of $5 \times 5$ convolutions. The DDU-Net architecture is not restricted to this specific network configuration. Adjustments such as altering the number of layers, dilation, and kernel size may enhance the model's receptive field size for specific applications. We choose our communication network to be fully convolutional to ensure adaptability to arbitrary input sizes as well as scaling the number of subimages.

The communication network receives as input $F$ feature maps from the deepest encoder layers for each of the $N \times M$ subimages; cf.~\cref{subsub:subnetwork}. These feature maps are concatenated along the height and width dimensions, aligning with the subimages' positions in the full-resolution input image; cf.~\Cref{fig:proposed-architecture}. The communication network processes this concatenated input and produces output feature maps with the same spatial dimensions as the inputs. These feature maps are partitioned back along height and width into $N \times M$ sub-feature maps, which are then sent back to the corresponding encoder-decoder network for further processing; they replace the original feature maps generated by the encoder and used as inputs to the communication network.

\subsection{Notation}
\label{subsec:notation}
In the remainder of this paper, we compare various model configurations using a naming convention for our domain decomposition-based U-Net that encapsulates key parameters. Each model is denoted as \DDUNet{D}{F}{C}, where:

\begin{itemize}
    \item $D$ (\(\in \mathbb{N}_{>0}\)) indicates the number of up- and down-sampling blocks in the encoder-decoder network.
    \item $F$ (\(\in \mathbb{N}_{\geq0}\)) specifies the number of feature maps processed by the communication network.
    \item $C$ (Y or N) denotes whether communication between communication feature maps is enabled ($C=Y$) or disabled ($C=N$)).
\end{itemize}
Note that, if communication is disabled ($C =$ N) but the number $F$ of feature maps processed by the communication network is non-zero, this means that $F$ feature maps are sent through the communication network independently, without concatenation. This effectively results in a baseline U-Net architecture with extra convolutions in the bottleneck layer.

For instance, \DDUNet{4}{64}{Y} refers to a model where the encoder-decoder network has a depth of $4$, $64$ feature maps are sent to the communication network, and communication is enabled. The case \DDUNet{D}{0}{N} operating on \(1 \times 1\) subimages is equivalent to the baseline U-Net architecture. \DDUNet{D}{F}{N}, with $F>0$ refers to a baseline U-Net with extra convolutions in the bottleneck layer. Additionally, for the case where $F = 0$, it does not matter whether communication is enabled (Y) or not (N), as the communication network uses $0$ feature maps.

\subsection{Model training}
\textbf{Loss Function.} For the network training, we employ the dice loss function, as this function addresses the issue of class imbalance for semantic segmentation of images; cf.~\cite{ma2021loss, ShrutiSurveyofLossFunctions}. The dice loss is defined as
\begin{equation} \label{eq:dice-loss}
	DL = 1 - 2 \frac{\sum_{p=1}^P \sum_{k=1}^K y_{k,p} \hat{y}_{k,p} + \epsilon}{\sum_{p=1}^P\sum_{k=1}^K y_{k} + \sum_{p=1}^P\sum_{k=1}^K \hat{y}_{k} + \epsilon}.
\end{equation}
Here, $K$ represents the number of classes, $P$ is the total number of pixels in a batch, and $\hat{y}_{k,p}$ is the predicted probability of pixel $p$ for class $k$, obtained by applying a softmax function to the model's output so that all the output logits are in the range $[0,1]$. Furthermore, $y_{k,p}$ denotes the true probability of pixel $p$ for class $k$, with values restricted to $\{0,1\}$ as the true mask is known, and $\epsilon$ serves as a small numerical stability constant (to avoid division by zero), set to $\epsilon=10^{-7}$ in this paper. 

\textbf{Parameter optimization}. The weights of the network are initialized using the method proposed by He \textit{et al.} \cite{he2015delving}. We employ the Adam optimizer~\cite{kingma2014adam}, with momentum parameters $\beta_1 = 0.9, \beta_2 = 0.999$, and a plateau learning rate decay strategy, where the learning rate decays with a factor of $2$ when the validation loss does not decrease, maintaining a patience of a chosen number of epochs. The initial learning rate, batch size, and early stopping criterion were chosen depending on the task and available memory and will be provided in the results section.

\textbf{Training procedure.} For training the DDU-Net architecture, we first initialize the encoder-decoder network with shared weights and distribute clones of this network  across all $M \times N$ devices. The coarse network for communication is initialized on a selected device and, due to the small size of the communicated feature maps, may optionally reside on the same device as one of the encoder-decoder networks. During forward propagation, each device independently computes the encoded feature maps. After processing a selected number $F$ of those feature maps using the communication network, the decoding of the feature maps again takes place fully in parallel. Backward propagation through the encoders and decoders can, again, be done in parallel without dependencies between the local encoder-decoder networks. However, after backward propagation through the decoders, communication is necessary for backward propagating through the communication network. After backward propagation, gradients are accumulated centrally on a main device and used to update weights, ensuring uniform updates across all devices. The updated weights are broadcasted to the other devices for synchronization.

\textbf{Implementation.} The implementation was done using PyTorch~\cite{pytorch} (version \texttt{1.12.0}), an open-source machine learning library.  We carried out the training and testing on the DelftBlue supercomputer~\cite{DHPC2024} at the Delft University of Technology, and we employed NVIDIA Tesla V100S GPUs, with a memory of $32$ GB, for the training.

\subsection{Datasets}
For testing our model, we use two different image datasets: 1) a synthetically generated dataset designed to test the capabilities of the communication network and 2) a realistic image semantic segmentation dataset for multi-class land cover segmentation to assess the effectiveness of the proposed model both in terms of segmentation quality and memory efficiency. 

\subsubsection{Synthetic dataset}
\label{subsec:synthetic-dataset}
\begin{figure}[ht!]
    \centering
    \includegraphics[width=1.0\linewidth]{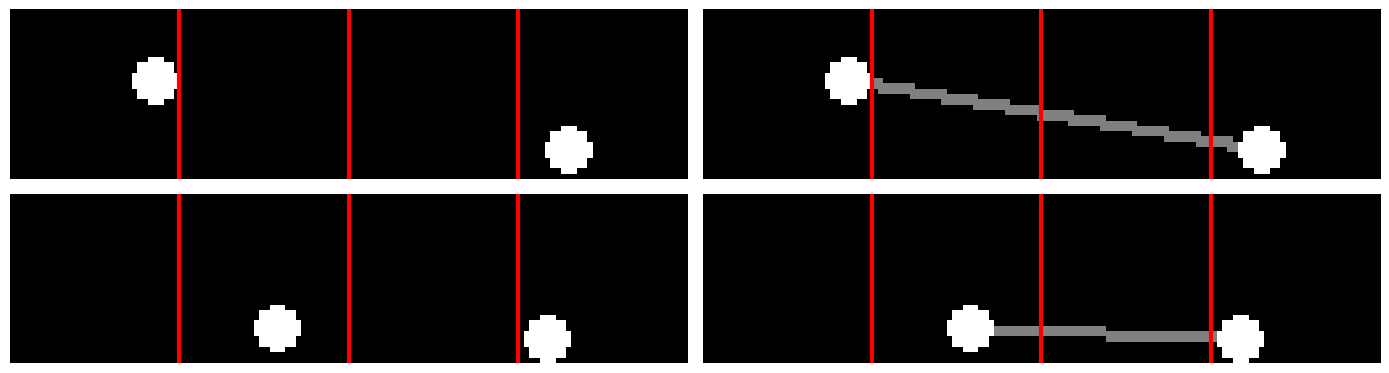}
    \caption{Two example images (left) and masks (right) from the synthetic dataset. The subimage boundaries used for these images are shown by the red vertical lines.}
    \label{fig:synthetic-example}
\end{figure}

In our approach, only deep feature maps, with low spatial resolution, are exchanged between encoder-decoder networks. This differs significantly from previous U-Net parallelization methods that involve the exchange of a number of feature maps at each U-Net layer. To assess the level of spatial context that these low-resolution feature maps can capture, we designed a synthetic dataset. This dataset comprises one-channel gray-scale images with dimensions $(k \cdot 32) \times 32$ pixels, where $k \in \{2, 3, 4, 6, 8, 16\}$. For each $k$, we generate $4\,000$ training, $1\,000$ testing and $1\,000$ validation images, resulting in a total synthetic dataset size of $36\,000$ images. This design allows the decomposition of images into $k$ subimages of $32\times32$ pixels; see~\Cref{fig:synthetic-example} for examples with $k=4$ subimages.

Each image in this gray-scale dataset displays a black background with two randomly placed white circles, each with a 4-pixel radius, placed completely within a subimage such that the subimage boundaries do not intersect the circles. The corresponding mask resembles the image, except for the addition of a third class of pixels: a line segment drawn between the centers of the two circles. Two examples of images and corresponding masks are shown in~\Cref{fig:synthetic-example}.

When processing an image in this dataset with the DDU-Net, the segmentation of the line segment connecting the two circles relies entirely on the communication network. As the two subimages are processed independently and only linked by the communication network, the effectiveness of the segmentation is a direct reflection of the communication network's capability to transfer global information accurately. 

\subsubsection{DeepGlobe land cover classification dataset} \label{sububsec:deep_globe}
\begin{figure}[ht!]
    \centering
    \includegraphics[width=1.0\linewidth]{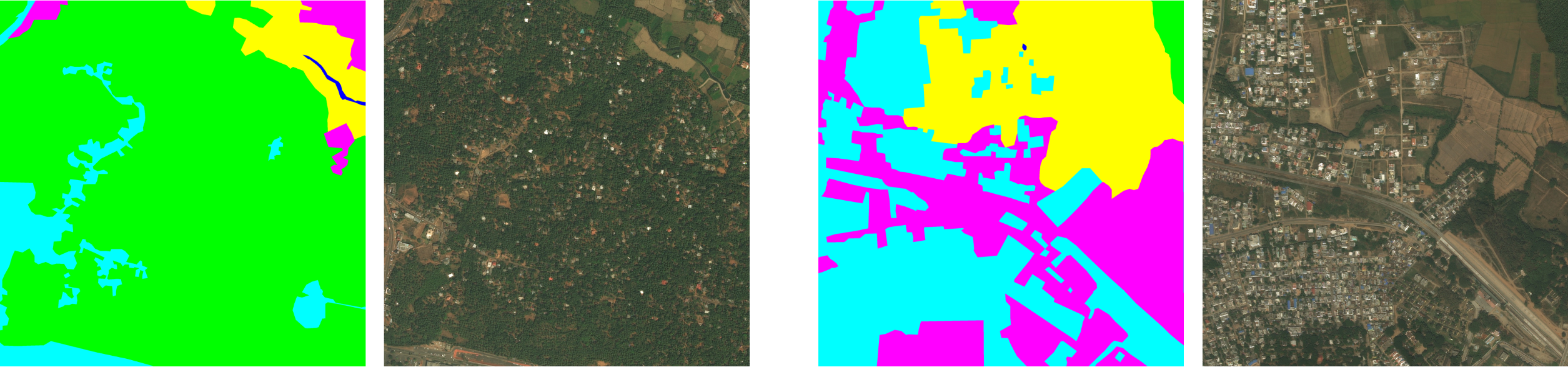}
    \caption{Two example images (right) and their corresponding masks (left) from the DeepGlobe land cover classification dataset~\cite{DeepGlobe18}.}
    \label{fig:deepglobe-example}
\end{figure}

The DeepGlobe land cover classification dataset~\cite{DeepGlobe18} is a semantic segmentation dataset for land cover types. The dataset contains 803 high-resolution ($2\,448 \times 2\,448$ pixels) annotated satellite images with 7 classes: urban, agriculture, rangeland, forest, water, barren, and unknown. The images have $50$\,cm/pixel resolution and span a total area of $1\,716.9$\,km$^2$. In addition to the high resolution of the images, segmenting this dataset is challenging due to the large class imbalance; see~\cref{tab:class-distributions-deepglobe}. In~\cref{fig:deepglobe-example}, we show two example images and masks from the dataset.

Other challenges include the limited number of images, inexact ground truth, and the presence of multi-scale relevant contexts; cf.~\cite{liu2020dense, khan2021deep}. To illustrate this, note that trees can exist in various land types such as urban, rangeland, or forest areas. As a result, the network needs to assimilate contextual information from a wider region around the tree to predict the correct class for the tree pixels accurately.

\begin{table}[ht!]
\centering
\begin{tabular}{lrr}
	\hline
	\textbf{class} & \textbf{pixel count} & \textbf{proportion} \\ \hline
	urban          &               642.4\,M &              9.35\,\% \\
	agriculture    &              3898.0\,M &             56.76\,\% \\
	rangeland      &               701.1\,M &             10.21\,\% \\
	forest         &               944.4\,M &             13.75\,\% \\
	water          &               256.9\,M &              3.74\,\% \\
	barren         &               421.8\,M &              6.14\,\% \\
	unknown        &                 3.0\,M &              0.04\,\% \\ \hline
\end{tabular}
\caption{Class distributions in the DeepGlobe land cover classification dataset. Table retrieved from~\cite{DeepGlobe18}.}
\label{tab:class-distributions-deepglobe}
\end{table}

\subsection{Evaluation Metrics}
After training the model, we analyze the results. Because of the large class imbalance, we used the (mean) class-wise intersection over union (IoU) score as a metric, as was suggested in the paper introducing the DeepGlobe dataset~\cite{DeepGlobe18}. Given a dataset with $n$ images, the IoU score for class $j$ is defined as: 
\begin{equation} \label{eq:class-iou}
    IoU_j = \frac{\sum_{i=1}^n TP_{ij}}{\sum_{i=1}^n TP_{ij} + \sum_{i=1}^n FP_{ij} + \sum_{i=1}^n FN_{ij}},
\end{equation}
where:
\begin{itemize}
    \item $TP_{ij}$ is the number of pixels in the $i$-th image correctly predicted as class $j$,
    \item $FP_{ij}$ is the number of pixels in the $i$-th image incorrectly predicted as class $j$, and
    \item $FN_{ij}$ is the number of pixels in the $i$-th image that belong to class $j$ but were predicted as another class.
\end{itemize} 
The mean IoU score ($mIoU$) provides a single metric by averaging the IoU scores across all classes, and is then defined as:
\begin{equation}
    mIoU = \frac{1}{K}\sum_{j=1}^K IoU_j,
\end{equation}
where $K$ is the number of classes.

\section{Architecture Discussion}
\label{sec:architecture}
In this section, we analyze some architectural properties of the DDU-Net.  First, we investigate the memory requirements of our approach both experimentally and theoretically and compare the results against a standard U-Net model. Then, we analyze the size of the receptive field of DDU-Net models with different architectures, that is, with varying depth of the subnetworks and the communication network.

\subsection{Memory requirements} \label{subsec:memory}
We compare the memory requirements for the baseline U-Net depicted in~\cref{fig:unet} and the proposed DDU-Net model architecture, both with a depth of $4$ up-sampling and down-sampling blocks, operating on 2 subimages; cf.~\cref{sec:methodology,fig:proposed-architecture,fig:proposed_subnetwork,fig:proposed_communication_network}. We specifically present results for a configuration with $F=64$ communicated feature maps. The channel distribution for the encoder-decoder networks in both models follows the same scheme depicted in~\cref{fig:unet}.

\Cref{tab:memory-and-weight-analysis} presents a detailed analysis of the memory requirements for the encoder and decoder of the U-Net architecture, which are identical to the encoder and decoder used in the DDU-Net, during training on a $1\,024 \times 1\,024$ image. The table also includes the memory requirements for the proposed communication network with $F=64$. The values in this table were derived theoretically (see, for instance,~\cite{sewak2018practical}) and validated experimentally using the \texttt{torch} library \cite{pytorch}. 

From the analysis, it is evident that storing the feature maps demands significantly more memory than the storing of model weights. The shallow layers, including the input block, the first decoder block, and the last decoder block, collectively contribute to nearly half of the total memory allocation for feature maps. Conversely, the number of weights increases for the deeper layers of the U-Net due to the higher number of kernels and larger kernel sizes in these blocks. The coarse feature maps used in the communicated layer result in relatively low memory requirements for the communication network, demonstrating the efficiency of the proposed DDU-Net in terms of memory utilization.

\begin{figure}[ht]
    \centering
    \includegraphics[width=\linewidth]{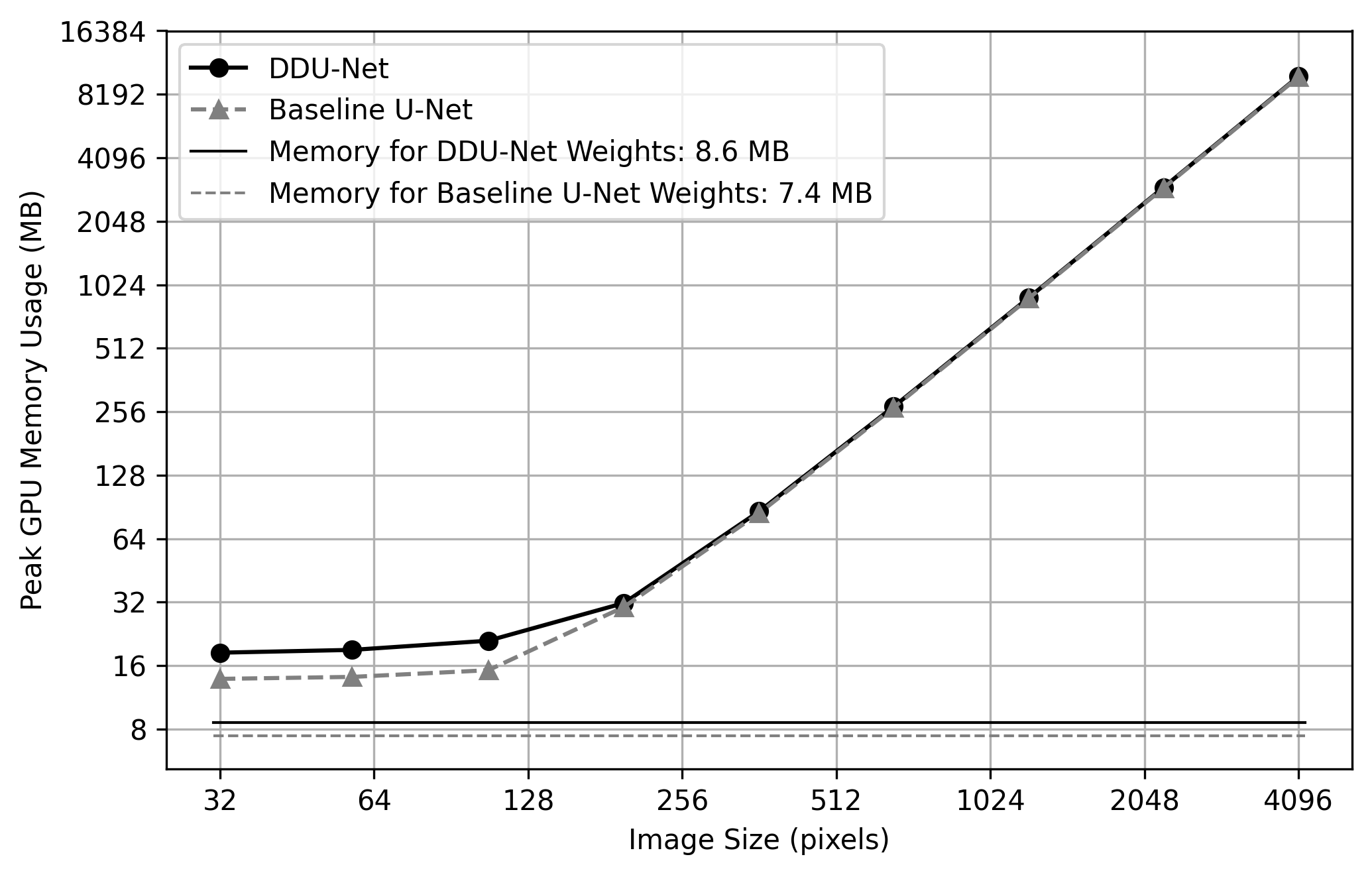}
    \caption{Measured peak memory of the proposed \DDUNet{4}{64}{Y} evaluated on two sub-images, compared to the baseline U-Net (with depth 4) evaluated on a single subimage, for various subimage resolutions during inference. The lines with markers represent the peak total memory usage for both networks, while the two horizontal lines only indicate the part of memory used for storing model weights. The experimental peak memory was measured using \texttt{torch}. For the DDU-Net, the GPU contains both the communication network (processing $F=64$ feature maps) and an encoder-decoder network. It is important to note that the U-Net peak memory usage is measured on a single GPU, while the DDU-Net peak memory usage is the maximum of the peak memory usage across two GPUs.
    }
    \label{fig:peak-memory-unet}
\end{figure}

Figure~\ref{fig:peak-memory-unet} shows the peak memory usage during inference for the baseline U-Net architecture and for a GPU containing both the U-Net encoder-decoder and the proposed communication network, for various image resolutions. For smaller image resolutions, the memory allocation can be largely attributed to model weights. However, as image size surpasses a moderate size of $2^7 \times 2^7$ ($128 \times 128$) pixels, the memory required to store the feature maps begins to dominate the total memory cost. Beyond this point, a noticeable increase in memory requirements is observed, as illustrated in \cref{fig:peak-memory-unet}. 
It is evident from~\cref{fig:peak-memory-unet} that the memory requirements for the DDU-Net and U-Net become relatively significantly closer at higher resolutions: for a resolution of $32$ pixels, the relative difference is $25.9\%$, whereas for $4\,096$ pixels, this difference reduces to only $0.65 \%$. Additionally, it is important to note that peak memory scales quadratically with resolution; therefore, doubling the resolution results in a fourfold increase in memory consumption.
The results show that the proposed DDU-Net, with its communication network, requires only a small memory overhead compared to the baseline U-Net, despite the added cost of the communication network. This demonstrates the DDU-Net's efficiency in memory utilization: it facilitates communication between sub-images while increasing the memory requirements only slightly.

\subsection{Receptive field analysis} \label{subsec:rf}
The proposed communication network operates on the coarse bottleneck layer of the encoder-decoder networks. Outputs of those layers typically have large receptive fields, as they are the result of numerous convolutional and downsampling operations. Therefore, the communication network is very effective in increasing the receptive field size of the model architecture. In~\cref{tab:receptive-fields}, we compare the (theoretical) receptive field size of a standard U-Net with that of our proposed model at different depths of the encoder-decoder network, following the theoretical approach as presented in \cite{richter2022receptive}. For the DDU-Net, we calculate the theoretical receptive field size for one infinitely large subdomain, providing an upper boundary for the true receptive field size when the network operates on multiple subimages. The comparison shows that the communication network significantly enlarges the receptive field size of the encoder-decoder network. 

\begin{table}[htbp]
\centering
\begin{tabular}{@{}cccc@{}}
\toprule
\textit{model / depth $D$ }& 2 & 3 & 4 \\ \midrule
baseline U-Net & $44 \times 44$ & $92 \times 92$ & $188 \times 188$ \\
DDU-Net({D},{F},{Y}) & $92 \times 92$ & $188 \times 188$ & $380 \times 380$ \\ \bottomrule
\end{tabular}
\caption[Receptive field size for different model depths]{Receptive field size for different encoder-decoder depths for baseline U-Net (cf.~\cref{fig:unet}) the DDU-Net (cf.~\cref{fig:proposed-architecture,fig:proposed_subnetwork,fig:proposed_communication_network}). Theoretical analysis of the receptive field was done following the approach described in~\cite{richter2022receptive}. Note that the number of feature maps  $F$ sent to the coarse network does not influence the receptive field size as long as $F>0$.
}
\label{tab:receptive-fields}
\end{table}

\section{Experimental Results}
\label{sec:experimental-results}
In this section, we compare the DDU-Net approach against the corresponding baseline U-Net model based on the segmentation quality on the datasets introduced in~\cref{sec:methodology}. Moreover, we conduct an ablation study to examine the effectiveness of the communication between encoder-decoder networks processing subimages.

\begin{table*}[ht]
\centering
\begin{tabular}{lrrrrrrr}
\hline
\bfseries name &
  \bfseries size &
  \begin{tabular}[c]{@{}r@{}} \bfseries $\#$ of input \\ \bfseries channels\end{tabular} &
  \begin{tabular}[c]{@{}r@{}} \bfseries $\#$ of output \\ \bfseries channels\end{tabular} &
  \begin{tabular}[c]{@{}r@{}} \bfseries memory feature \\ \bfseries maps ($\#$ of values)\end{tabular} &
  \begin{tabular}[c]{@{}r@{}} \bfseries memory feature \\ \bfseries maps (MB)\end{tabular} &
  \begin{tabular}[c]{@{}r@{}} \bfseries memory weights\\ \bfseries ($\#$ of values)\end{tabular} &
  \begin{tabular}[c]{@{}r@{}} \bfseries memory \\ \bfseries weights (MB)\end{tabular} \\ \hline
input                                                           & 1\,024 & 3     & 3    & 3.1\,M    & 12,0      & -             & 0.00       \\ \hline
input block                                                     & 1\,024 & 3     & 16   & 67\,M     & 256.0     & 2\,800        & 0.01 \\
encoder block 1                                                 & 512   & 16    & 32    & 42\,M     & 176.0     & 13\,952       & 0.05 \\ 
encoder block 2                                                 & 256   & 32   & 64     & 21\,M     & 88.0      & 55\,552       & 0.21 \\ 
encoder block 3                                                 & 128   & 64   & 128    & 10\,M     & 44.0      & 221\,696      & 0.85 \\ 
encoder block 4                                                 & 64    & 128   & 256   & 5.2\,M    & 22.0      & 885\,760      & 3.38 \\ \hline
\begin{tabular}[c]{@{}l@{}}
	communication \\
	   network
\end{tabular} & 64    & 64   & 64     & 7.6\,M    & 29.0      & 307\,776      & 0.04\\ \hline
decoder block 1                                                 & 64    & 256 & 128     & 13\,M     & 48.0      & 573\,952      & 2.19 \\
decoder block 2                                                 & 128   & 128   & 64    & 25\,M     & 96.0      & 143\,616      & 0.55 \\
decoder block 3                                                 & 256   & 64   & 32     & 50\,M     & 192.0     & 35\,968       & 0.14 \\
decoder block 4                                                 & 512   & 32   & 16     & 101\,M    & 384.0     & 9\,024        & 0.03 \\
output block                                                    & 1\,024 & 16    & 3    & 3.1\,M    & 12.0      & 51            & 0.00 \\ \hline 
labels                                                          & 1\,024 & 3     & 3    & 1.0\,M    & 8.0       & -             & - \\ \hline
\end{tabular}
\caption{Theoretical analysis of the memory requirements and number of weights of a 4-blocks deep U-Net encoder and decoder architecture (see~\cref{subsec:the-u-net}) training for an RGB image of $1\,024 \times 1\,024$ pixels and a 3-class segmentation task. The table displays the size, number of input and output channels, memory usage (in terms of number of values and megabytes), as well as weight count and size (in terms of values and megabytes) for each block of the encoder and decoder part of the CNN architecture as well as the communication network. The communication network memory is based on 4 subimages and communication of 64 feature maps per subimage. Note that the number of weights and their memory cost is independent of the image size, whereas the number of values and memory of the feature maps will increase as image size increases.}
\label{tab:memory-and-weight-analysis}
\end{table*}

\subsection{Synthetic Dataset Results}
In this section, we present the results obtained using our synthetic dataset, which is described in~\cref{subsec:synthetic-dataset}. The dataset comprises gray-scale images of varying dimensions, specifically $(k \cdot 32) \times 32$ pixels, where $k \in \{2, 3, 4, 6, 8, 16\}$. For each $k$, we trained separate baseline U-Nets and DDU-Nets.

For our experiments, we trained baseline U-Nets using entire $ k \cdot 32 \times 32$ images as inputs. In contrast, the DDU-Net architecture received $k$ subimages of size $32 \times 32$ pixels, processed by $k$ separate encoder-decoder clones. Both the U-Net and DDU-Net models were designed for 1-channel grayscale input images, generating outputs with three channels corresponding to the logits for the pixel classes: \textit{background}, \textit{line segment}, and \textit{circle}.

We evaluate  
\begin{itemize}
	\item the impact of varying the number of down- and up-sampling blocks (depth $D$) in the encoder-decoder networks, 
	\item the influence of the depth of the communication network, which is equal to the number of feature maps sent to the communication network (denoted as $F$), and 
	\item the effect of inter-feature map communication ($C$, with values $Y$ for enabled and $N$ for disabled).
\end{itemize}
As mentioned in~\cref{subsec:notation}, we use the notation \DDUNet{D}{F}{C} to represent a DDU-Net with parameters $D \in \mathbb{N}_{\geq 0}$, $F \in \mathbb{N}_{\geq 0}$, and $C \in \{Y, N\}$. For a baseline U-Net with depth $D$ we use the notation \UNet{D}. The training hyperparameters used to train the networks are listed in~\cref{tab:hyperparams1}. The number of weights in the encoder, decoder, and communication network for the different experiments can be found in the appendix, specifically in~\cref{tab:subnets-synthetic}.

\subsubsection{Qualitative results}
\begin{figure}[ht]
\centering
\includegraphics[width=\linewidth]{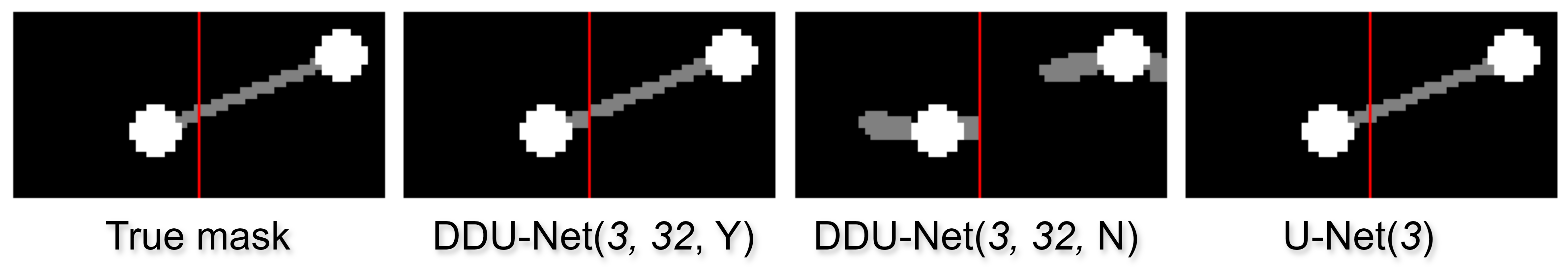}
\caption{Predictions for different models. The DDU-Net operates on two subimages, whereas the U-Net is trained on the full $64\times32$ image. We remind the reader here that the {\DDUNet{3}{32}{N}} processes the subdomains independently without communication (as indicated by the ``N''), whereas the {\DDUNet{3}{32}{Y}} includes communication. The red vertical line in the horizontal center of the image indicates the subimage border.
}
\label{fig:example-preds-synthetic-dataset}
\end{figure}

To show the effectiveness of the coarse network with communication, we compare the qualitative segmentation masks for three architectures: the \UNet{3}, the \DDUNet{3}{32}{Y} and the \DDUNet{3}{32}{N} for the $32 \times 64$ pixel images ($k=2$) in~\cref{fig:example-preds-synthetic-dataset}. We observe that the \DDUNet{3}{32}{N} is unable to predict the position of the line segment correctly, as it is unaware of the circle position in the other subimage due to the lack of communication. In contrast, the \DDUNet{3}{32}{Y} predicts the location of the line segment correctly.

\Cref{fig:example-preds-synthetic-dataset} shows that the proposed encoder-decoder network can capture the relevant features in the $F=32$ communicated feature maps, and that the communication network is able to transfer this information between the encoder-decoder networks. While we currently communicate all $32$ feature maps in the bottleneck of the $\DDUNet{3}{32}{Y}$ for this example, we will examine the impact of varying the number of feature maps in~\cref{subsubsec:varying-number-of-fmaps}.  This is particularly noteworthy, as the communication only happens on a very coarse level (with $4\times4$ pixel feature maps) and the line segments are drawn on the finer grained $64\times 32$ resolution. Apparently, communication on this coarse level is sufficient for the decoder to produce a good segmentation result on the fine level.

\subsubsection{Varying the number of subimages and communicated feature maps}
\label{subsubsec:varying-number-of-fmaps}
\begin{figure}[ht]
    \centering
    \includegraphics[width=0.8\linewidth]{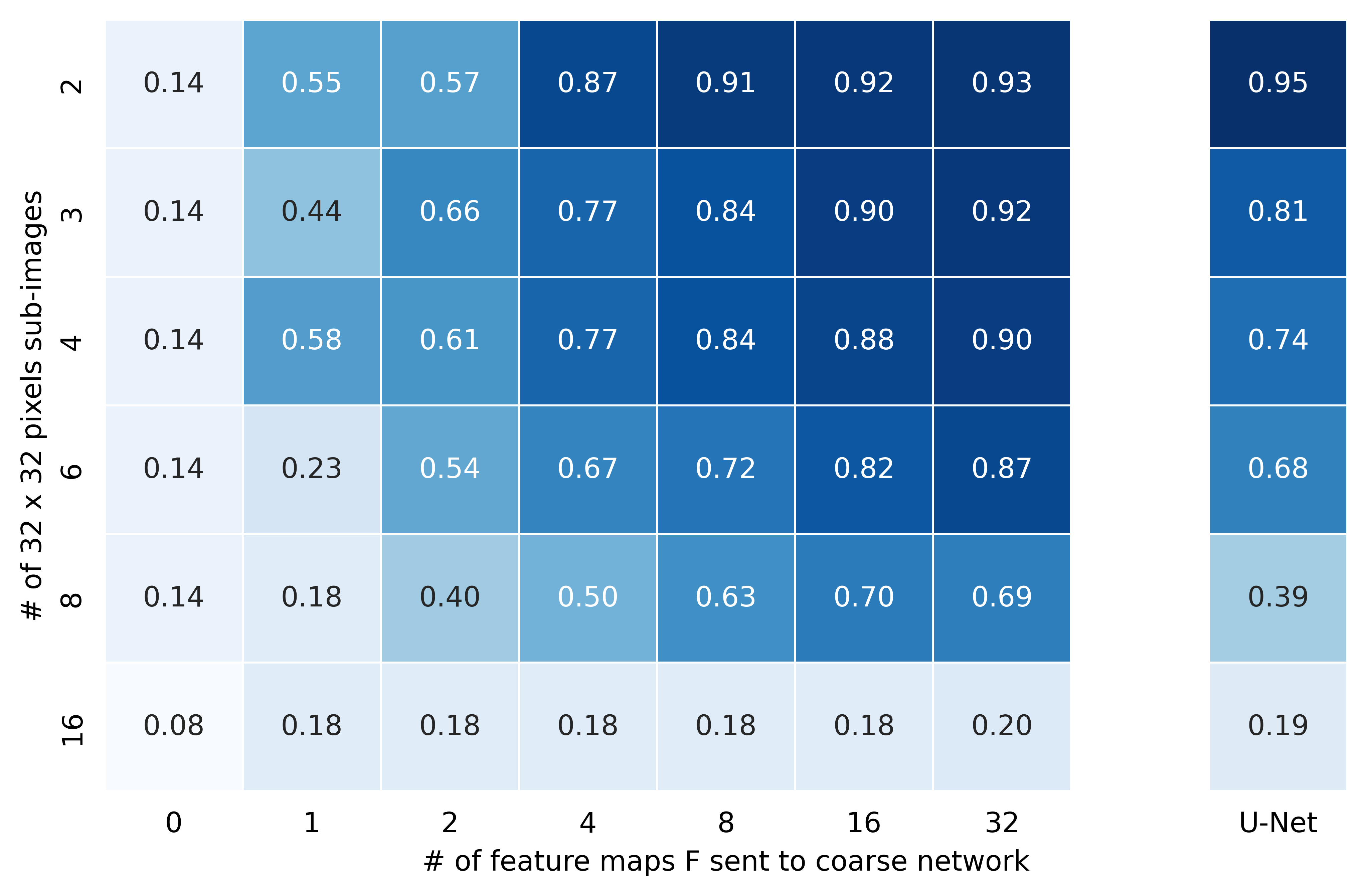}
    \caption{IoU score for the line segment class for a \UNet{3} and \DDUNet{3}{F}{Y} for different numbers of feature maps $F$ in the communication network and different image dimensions. Note the increase in IoU score when the number of communicated feature maps $F$ increases (horizontal).    
    }
    \label{fig:iou-scores-for-different-nums-of-subdomains}
\end{figure}

In~\cref{fig:iou-scores-for-different-nums-of-subdomains}, the IoU score (\cref{eq:class-iou}) for the line segment class is depicted for the baseline \UNet{3} as well as the \DDUNet{3}{F}{Y}, with $F\in \{0, 1, 2, 4, 8, 16, 32\}$ feature maps communicated, for different numbers of subimages $k$. Recall that the case $F=0$ corresponds to a DDU-Net where the communication network has a depth of $0$, meaning it is not used, also implying that there is no communication between encoder-decoder clones.
We observe two phenomena. First, there exists a clear positive correlation between the number of communicated feature maps $F$ and the quality of the results. Secondly, the DDU-Net performs even better than the baseline U-Net for larger images and numbers and subimages, respectively. This is related to the sizes of the receptive fields as reported in~\cref{tab:receptive-fields}: the DDU-Net has a larger receptive field ($188 \times 188$ pixels versus $92 \times 92$ pixels), yielding better results when the circles are further apart.

\subsubsection{Impact of the depth of the encoder-decoder network}
In the DDU-Net, the encoder-decoder network has depth $D$, that is, the number of up- and down-sampling blocks. The ideal depth is a trade-off: while shallow depths cannot capture all relevant features due to a limited receptive field size (cf.~\cref{tab:receptive-fields}), overly deep encoder-decoders contain more parameters and may produce too coarse-grained and less informative feature maps for communication.

\begin{figure}[ht]
    \centering
    
    \subfigure[$D=2$]{
        \includegraphics[width=0.95\linewidth]{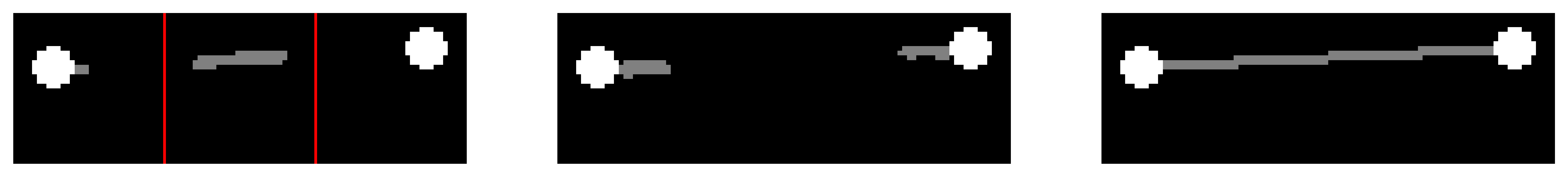}
    }\\[-0.1cm] 
    \subfigure[$D=3$]{
        \includegraphics[width=0.95\linewidth]{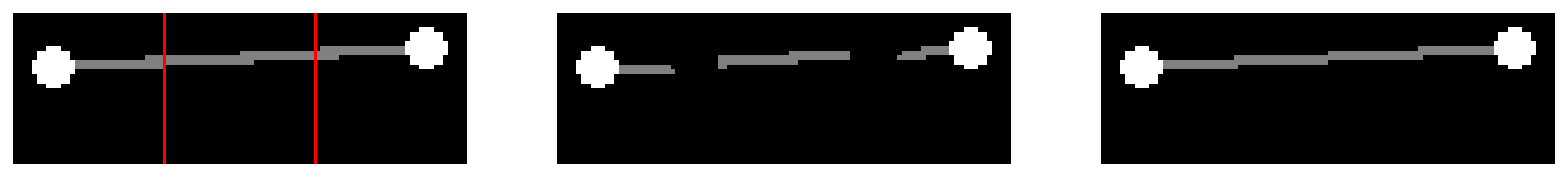}
    }\\[-0.1cm] 
    \subfigure[$D=4$]{
        \includegraphics[width=0.95\linewidth]{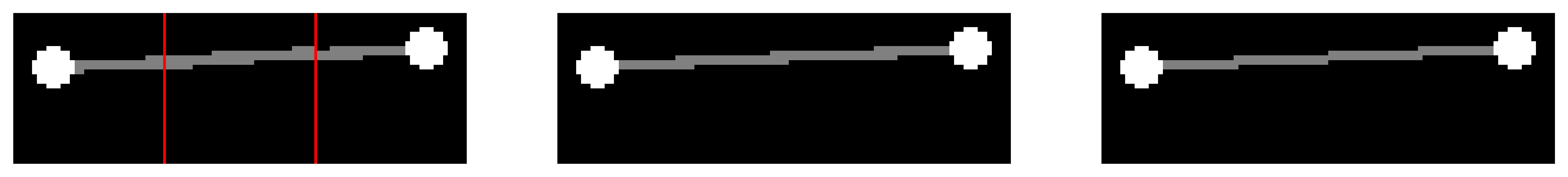}
    }
    \caption{From left to right: mask predicted by the \DDUNet{D}{16}{Y}, mask predicted with baseline \UNet{D}, and true mask for different encoder-decoder depths $D$. The subimage borders are indicated by the red vertical lines. Note that the baseline model is trained on the full image, so there are no subimages present for this case.  The broken lines for the baseline U-Net are caused by the limited receptive field size of this network.
    }
    \label{fig:examples-different-depths}
\end{figure}

To see the effect of the depth of the encoder-decoder on the predictions, we compare the segmentation masks generated by a \DDUNet{D}{16}{Y} to the segmentation masks generated by a \UNet{D}, with $D \in \{2,3,4\}$. The resulting predictions of those predictions are shown in~\cref{fig:examples-different-depths}. For encoder-decoder networks with $D=2$, we observe that the line segment is only predicted correctly for the pixels that are in the center region between the two circles. This can also be explained by the receptive field: shallow networks have a limited receptive field, such that pixels far away from one of the two circles are not affected by this circle; this leads to the incorrect segmentation result.
It can also be seen that the DDU-Net with communication enabled gives better results than the U-Net for $D=2$ and $D=3$, as the communication network increases the receptive field size; cf.~\cref{tab:receptive-fields}. 

\subsubsection{Generalization to different numbers of subimages}
\label{subsec:train-eval-on-different-numebrs}
\begin{figure}
    \centering
    \includegraphics[width=0.8\linewidth]{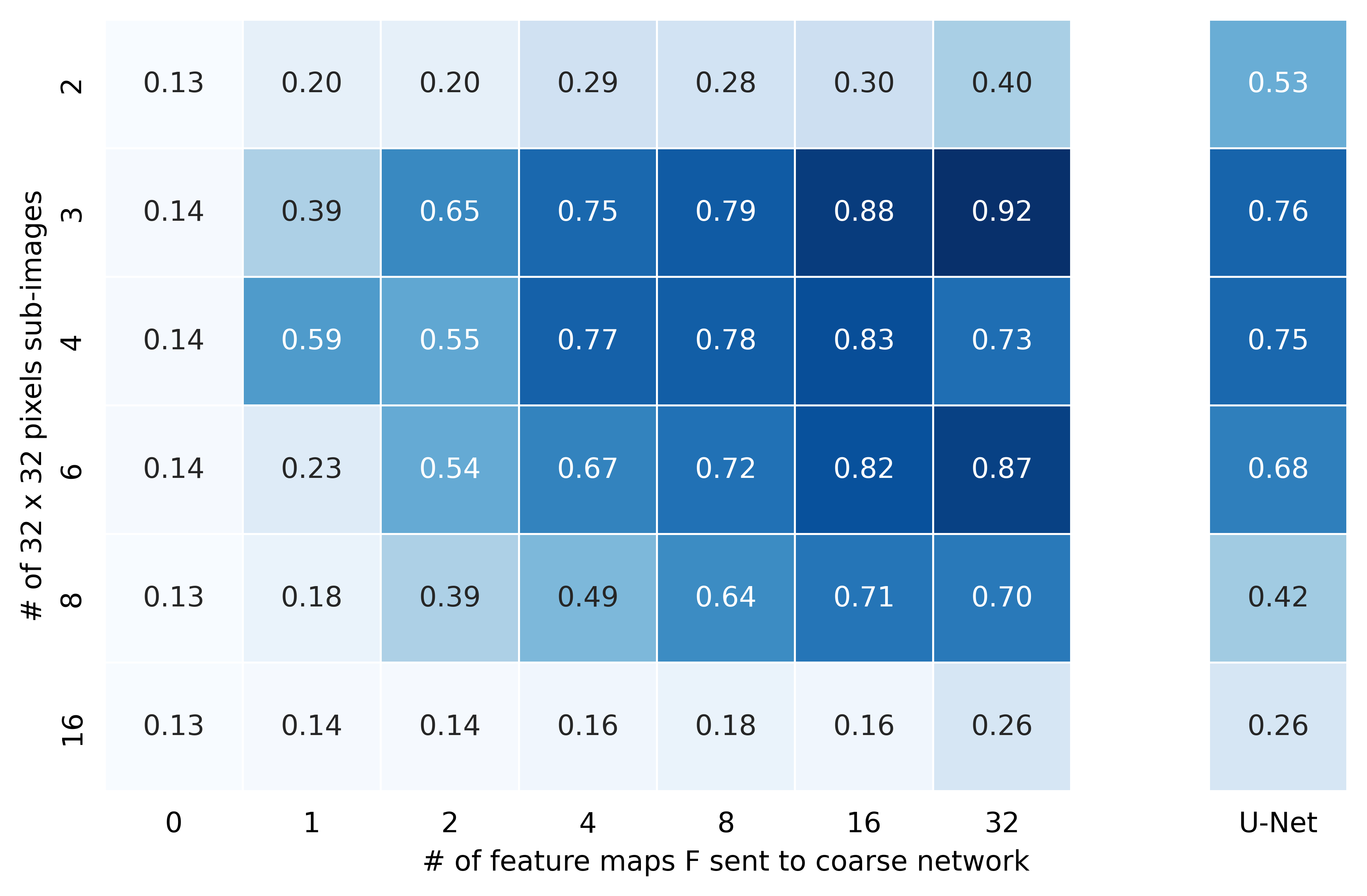}
    \caption{IoU score for the line segment class for a 3-deep encoder-decoder network in the DDU-Net for models trained on varying numbers of subimages with varying numbers and numbers of communicated feature maps. All models are evaluated on the same test dataset of $(6 \cdot 32) \times 32$ subimages.
    }
    \label{fig:iou-scores-same-test-dataset-different-train-datasets}
\end{figure}

The baseline U-Net and the DDU-Net are both size-agnostic, as both are fully convolutional neural networks. This allows them to process images with sizes different from the training images. In case of the DDU-Net, this means that we can vary both the size of subimages and the number of subimages. In order to test the generalization of the DDU-Net with respect to the number of subimages for different amounts of communication, we train for each $k\in\{2,3,4,6,8,16\}$ a \DDUNet{3}{F}{Y} on input images with spatial dimensions of $(k \cdot 32) \times 32$ pixels, as described in section~\cref{subsec:synthetic-dataset}, with $F \in \{0,1, 2,4,8,16,32\}$. Each of these models is then evaluated on a dataset with 6 subimages. The results are shown in~\cref{fig:iou-scores-same-test-dataset-different-train-datasets}.

We can make the following observations: 
\begin{enumerate}
\item Models trained on $3$, $4$, and $6$ subimages perform the best on the $6$ subimage test dataset. This is likely because the images in these datasets are small enough to be (almost) fully covered by the $188\times 188$ pixels receptive field of the DDU-Net with $D=3$; cf.~\ref{tab:receptive-fields}.
\item The more inaccurate results for the $8$ and $16$ subimages can be explained by the limited receptive field size, $188\times 188$ pixels, of the DDU-Net with $D = 3$; cf.~\cref{tab:receptive-fields}. 
\item The IoU score deteriorates when the model is trained on only two subimages and then evaluated on a larger number of subimages. This is likely due to the fact that every subimage in the training data set contains one circle and a part of the line segment; however, there are no samples with subimages that do not contain circles or line segments.
\end{enumerate}

These findings suggest that the DDU-Net can be trained on a different number of subimages than on which it is evaluated. 

\subsubsection{Summary of results on the synthetic dataset}
We summarize our findings for the synthetic dataset as follows:
\begin{itemize}
    \item The communication network is able to transfer contextual information across subimages, which becomes clear both from qualitative comparison in~\cref{fig:example-preds-synthetic-dataset} and quantitative results in~\cref{fig:iou-scores-for-different-nums-of-subdomains}.
    \item The segmentation quality increases when the number of feature maps involved in communication increases.
    \item The larger receptive field size of the DDU-Net with communication leads to better results compared to the baseline U-Net; cf.~\cref{fig:iou-scores-for-different-nums-of-subdomains,fig:examples-different-depths}.
    \item The DDU-Net can be trained successfully on a fixed number of subimages and evaluated on different number of subimages. 
\end{itemize}

\subsection{DeepGlobe Dataset}
Now, we assess the effectiveness of the DDU-Net for a real-world dataset with high-resolution images: the DeepGlobe dataset; cf.~\cref{sububsec:deep_globe}.

\begin{figure}[ht]
    \centering
    \includegraphics[width=\linewidth]{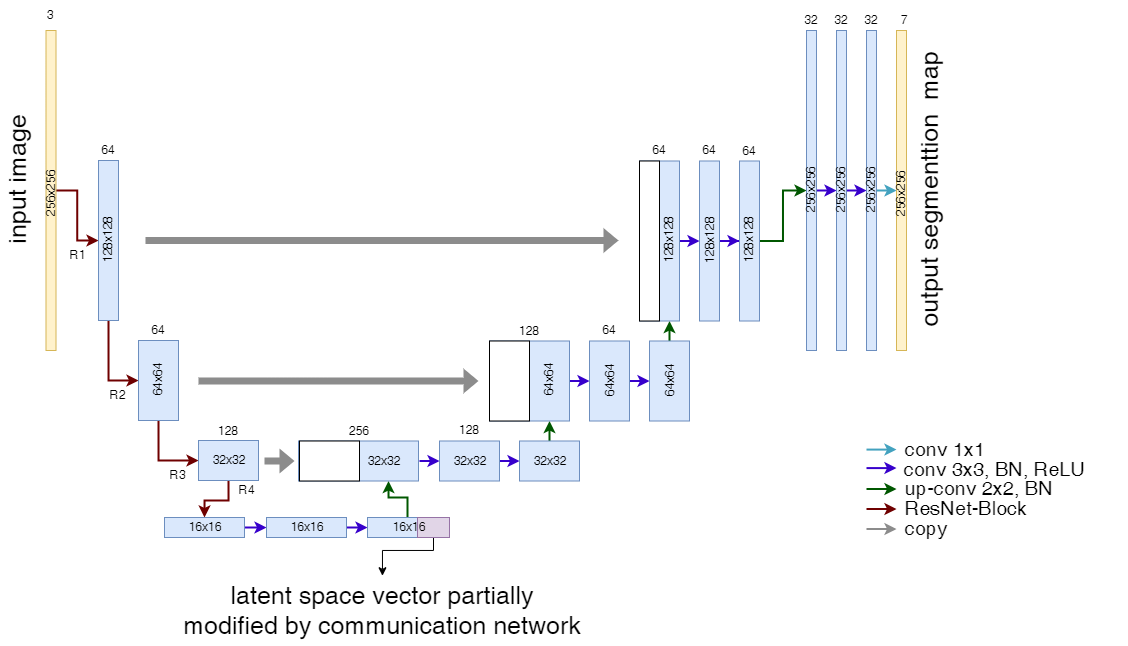}
    \caption{Architecture of the encoder-decoder ResNet-UNet achitecture for image segmentation. The pre-trained blocks of the ResNet-18 are shown as $R1, R2, R3, R4$. Note that the inner architecture of these blocks is not shown completely, but is largely simplified here.}
    \label{fig:resnet-unet}
\end{figure}

Given the small number of 803 images in the DeepGlobe dataset relative to the segmentation task's complexity, we made several adjustments to the encoder-decoder architecture, including batch normalization, random dropout layers, and data augmentation (horizontal and vertical flipping, random rotation of $R$ radians, with $R \in \{0, \pi/2, \pi, 3\pi/4\}$), and we incorporated a pre-trained image encoder model, the ResNet-18~\cite{he2016deep}. This model, trained on over $1\,000$ classes on the ImageNet data set \cite{deng2009imagenet}, offers a rich feature representation for diverse images. The ResNet-18 model consists of residual blocks with skip connections allowing an effective gradient flow. We used the first four residual blocks of the ResNet-18 to initialize our encoder. This strategy leverages the pre-existing knowledge within the pre-trained model to enhance the network's ability to generalize patterns from the limited dataset. A visualization of the employed model architecture is given in~\cref{fig:resnet-unet}. For a detailed overview of the distribution of model parameters within each component of the network, we refer to~\cref{tab:subnets-deepglobe}.
 
During training, we kept the ResNet-18 model's weights fixed, only adjusting the weights of the decoder and the communication network. Additionally, we inserted two extra $3\times3$ convolutional layers in the bottleneck layer. These layers allow the network to restructure and refine the feature maps produced by the ResNet encoder generate relevant information for the communication network.

To ensure a fair comparison between the segmentation quality of the baseline U-Net and the DDU-Net, we trained both models on equally large images. Due to the baseline U-Net architecture's inherent inability to parallelize across different GPUs and the limited training device memory (32 GB), we cropped $1\,024 \times 1\,024$ non-overlapping patches from the DeepGlobe dataset, resulting in a training dataset consisting of $2\,412$ images. It is crucial to distinguish between these large ``global patches'' and the subimages in the DDU-Net architecture. During training of the DDU-Net, the global $1\,024 \times 1\,024$ patches are further partitioned into smaller subimages, which then are distributed across the encoder-decoder networks. Using mixed precision training~\cite{micikevicius2017mixed}, this approach allowed us to train with a mini-batch size of 12 images on a single GPU. This and other hyperparameters used for training are shown in~\cref{tab:hyperparams-deepglobe}. Furthermore, to take into account random initialization of the network parameters, we trained every network repeatedly for three times with the same settings and training dataset. For each configuration, the best performing model (in terms of IoU score on the test dataset) was selected. 
\begin{table}[ht]
\centering
\begin{tabular}{@{}lr@{}}
\toprule
\multicolumn{2}{c}{\textbf{hyperparameters}} \\ \midrule
learning rate                   & 0.001      \\
number of epochs                & 100         \\
dropout rate                    & 0.1        \\
batch size                      & 12         \\ 
loss function                   & dice loss (\cref{eq:dice-loss}) \\
\bottomrule
\end{tabular}
\caption[Hyperparameters for training on DeepGlobe Satellite Dataset]{Hyperparameters used for training the model on the DeepGlobe land cover classification Dataset}
\label{tab:hyperparams-deepglobe}
\end{table}

We want to investigate if the DDU-Net can perform equivalently to the baseline U-Net. However, when we include communication in the DDU-Net architecture the additional trainable parameters imply a potential ability to learn more complex patterns. To isolate the effects of (1) the \textit{extra parameters} in the communication network and (2) the \textit{communication itself}, we vary the number of feature maps $F$ for both \DDUNet{4}{F}{Y} and \DDUNet{4}{F}{N}. The \DDUNet{4}{F}{N} scenario represents an encoder-decoder network with an extra coarse network at the bottleneck layer (with coarse network chosen the same as the communication network), but \textit{without} information exchange between sub-domains. In this case, the coarse network operates solely on the \textit{local} bottleneck feature maps, rather than on the concatenated feature maps of all subimages. Notably, $F=0$ corresponds to the baseline \UNet{4}. The difference between the cases with communication enabled ($Y$) and disabled ($N$) shows the impact of communication across subimages.

\subsubsection{Quantitative results}
\begin{figure}[ht]
    \centering
    \includegraphics[width=\linewidth]{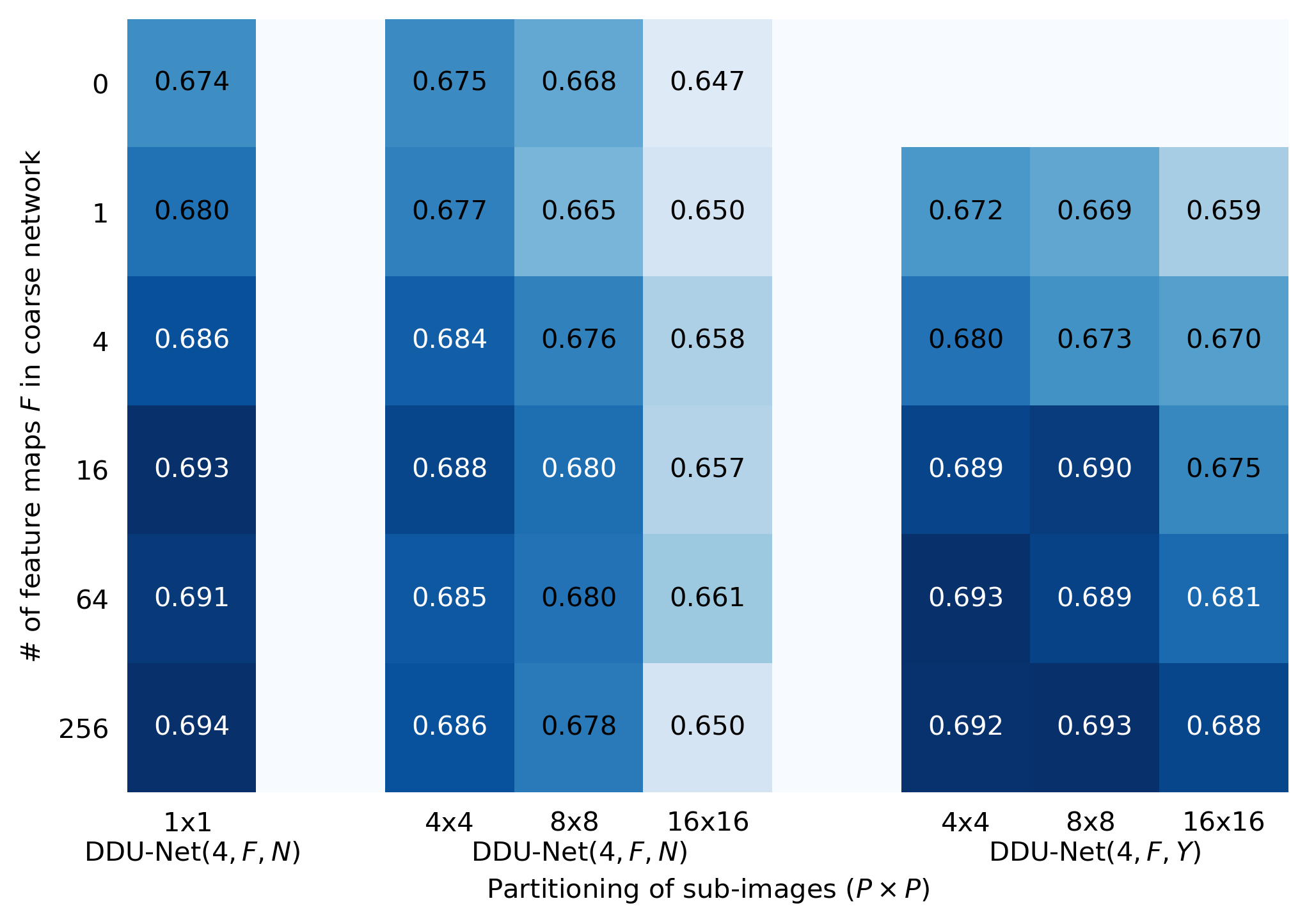}
    \caption{
    Mean IoU scores for various configurations of the DDU-Net architectures on a $2\,048 \times 2\,048$ test dataset. The left section presents results for a baseline \DDUNet{4}{F}{N} model evaluated on entire images ($P=1$). The middle and right sections show the performance of the \DDUNet{4}{F}{N} and \DDUNet{4}{F}{Y} models, respectively, trained on $P\times P$ subimages that form a non-overlapping partition of the full training image, with $P\in \{4,8,16\}$. The experiments vary the partitioning $P\times P$ of the subimages and the number of feature maps $F$ in the coarse network. Note that the baseline U-Net is the same model as the \DDUNet{4}{0}{N} evaluated on $1\times1$ subimages}.
    \label{fig:evaluation-results-deepglobe}
\end{figure}

\Cref{fig:evaluation-results-deepglobe} shows the mean IoU scores for different DDU-Net configurations and image partitions, trained on $1\,024\times 1\,024$ images and evaluated on the $2\,048 \times 2\,048$ test dataset with fixed subimage size, using the approach discussed in~\cref{subsec:train-eval-on-different-numebrs}. 

From this figure, two main observations emerge. First, the segmentation quality improves with increased number of feature maps $F$ in the coarse network. This is expected since a higher number of feature maps leads to an increase in the number of parameters and a larger receptive field, enabling the model to capture more complex and distant patterns.
Another trend in~\cref{fig:evaluation-results-deepglobe} is that the quality of the predictions decreases as the number $P\times P$ of subimages increases, or, equivalently, as the subimage size decreases. However, for the \DDUNet{4}{F}{Y}, this decrease in quality is much less pronounced compared to the \DDUNet{4}{F}{N}. This indicates that the communication combined with the coarse network effectively transfers contextual information between subimages, also for this realistic dataset. 

\begin{figure}[ht]
    \centering
    \includegraphics[width=1.0\linewidth]{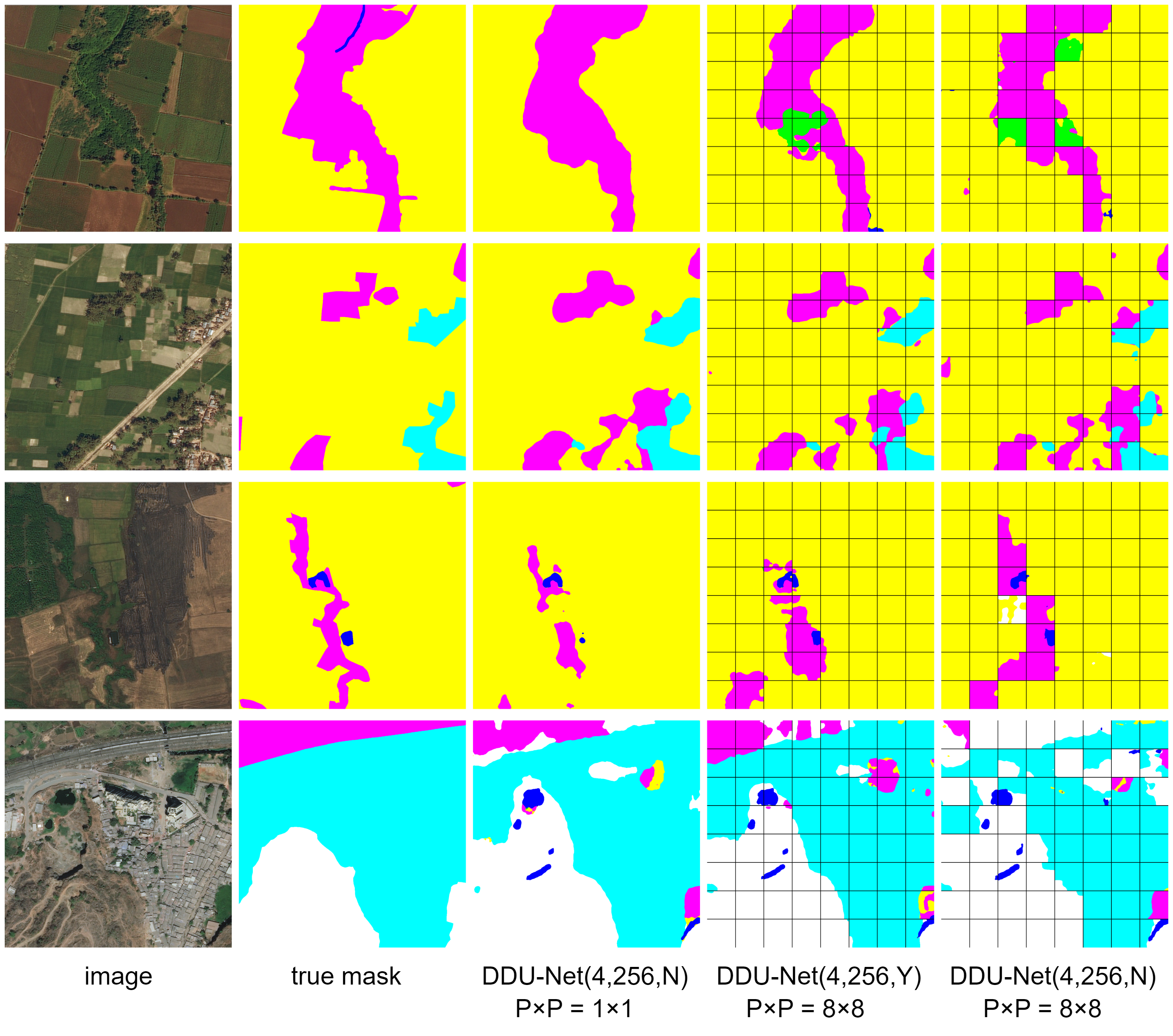}
    \caption{From left to right: original image, true mask, predictions by the \DDUNet{4}{256}{N} evaluated on the full image, predictions by the \DDUNet{4}{256}{Y} evaluated on $8 \times 8$ subimages, and predictions by the \DDUNet{4}{256}{N} also evaluated on $8 \times 8$ subimages. The borders of the $8 \times 8$ subimages are indicated by black lines. Prediction errors for these images are visualized in~\cref{fig:qualitative-deepglobe-predictions-differences}. Note that the \DDUNet{4}{256}{N} evaluated on the full image ($P\times P = 1\times 1$) is equivalent to a $5$-deep baseline U-Net.}\label{fig:qualitative-deepglobe-predictions}
\end{figure}

\subsubsection{Qualitative results}
In~\cref{fig:qualitative-deepglobe-predictions}, example predictions are shown for different training and model configurations. We observe that the DDU-Net trained on $1\times 1$ and the DDU-Net trained on $8 \times 8$ subimages with communication enabled both produce good segmentation results, although there are differences compared to the true mask. Conversely, the DDU-Net trained on $8\times 8$ subimages without communication shows poorer predictions.To better illustrate prediction errors, we include \cref{fig:qualitative-deepglobe-predictions-differences} in the Appendix, where black indicates correct predictions and white indicates incorrect predictions. 
The key distinction lies in the consistency across neighboring subimages: when communication is enabled, predictions for neighboring subimages exhibit smoother boundaries instead of patchy patterns. This highlights the effectiveness of communication between subimages in the DDU-Net architecture.

\subsection{Comparison to other methods}
We compared the DDU-Net with two high-resolution segmentation methods: GL-Net \mbox{\cite{chen2019collaborative}} and the From-Contexts-to-Locality (FCtL) network \mbox{\cite{li2021contexts}}. Unlike these methods, the DDU-Net adopts a fundamentally different approach. Existing methods focus on minimizing computational workload to fit tasks onto a single GPU, often sacrificing critical high-resolution details. In contrast, the DDU-Net distributes the workload across multiple GPUs, enabling efficient processing of large-scale, high-resolution data without losing detail. Future research could explore integrating the strengths of both approaches, balancing efficient workload distribution with single-GPU compatibility.

We trained a DDU-Net with ResNet50 and ResNet18 encoder backbones, the FCtL network, and GL-Net on the same DeepGlobe dataset of $2\,048 \times 2\,048$ images. DDU-Net training followed the settings in Table{~\ref{tab:hyperparams-deepglobe}}, while FCtL and GL-Net adhered to their respective papers' configurations.

The results, summarized in Table{~\ref{tab:comparison}}, compare accuracy, peak memory usage, and inference time on the DelftBlue supercomputer \mbox{\cite{DHPC2024}}. The DDU-Net is uniquely capable of parallelizing tasks across multiple GPUs. For instance, the DDU-NetResNet18 with $2 \times 2$ subdomains achieves a mean IoU of 0.684, requiring just 601 MB of memory and $0.0463 \pm 0.0011$ seconds per $2\,048 \times 2\,048$ image. In contrast, state-of-the-art methods like GL-Net and FCtL require over 2 GB of memory and inference times exceeding 3 seconds. 
A significant part of this performance gap stems from fundamental differences in approach. Unlike FCtL and GL-Net, which is designed to process parts of the image sequentially on a single GPU, the DDU-Net distributes the entire image workload across multiple GPUs, which enables more efficient parallelization. 
While the DDU-Net with a ResNet50 backbone uses more memory than FCtL and GL-Net, its memory usage scales almost linearly with the number of GPUs while maintaining a consistent mean IoU, which is exactly the aim of the DDU-Net.

\begin{table*}[t]
\centering
\resizebox{0.9\textwidth}{!}{%
\begin{tabular}{lcccc}
\hline
\multicolumn{1}{c}{\textbf{method (\# of GPUs)}} & \begin{tabular}[c]{@{}c@{}}\textbf{mean} \\ \textbf{IoU}\end{tabular} & \begin{tabular}[c]{@{}c@{}}\textbf{number} \\ \textbf{of parameters}\end{tabular} & \begin{tabular}[c]{@{}c@{}}\textbf{peak memory usage for a}\\  $2\,048 \times 2\,048$ \textbf{image (MB)}\end{tabular} & \begin{tabular}[c]{@{}c@{}}\textbf{inference time }\\ \textbf{average (s)}  $\pm$ \textbf{std (s)}\end{tabular} \\ \hline
DDU-NetResNet50 ($1\times 1$ subdomains, 1 GPU) & 0.704 & 44M & $3\,834$ & $0.2637 \pm 0.0023$ \\
DDU-NetResNet50 ($2\times 2$ subdomains, 4 GPUs) & 0.703 & 44M & $1\,109$ & $0.0983 \pm 0.0016$ \\ \hline
DDU-NetResNet18 ($1\times 1$ subdomains, 1 GPU) & 0.690 & 9.7M & $2\,742$ & $0.0857 \pm 0.0004$ \\
DDU-NetResNet18 ($2\times 2$ subdomains, 4 GPUs) & 0.684 & 9.7M & $739$ & $0.0463 \pm 0.0011$ \\ \hline
From Contexts to Locality \cite{li2021contexts} (1 GPU) & \textbf{0.705} & 54.1M & $3\,131$ & $3.3907 \pm 0.0150$ \\ \hline
GL-Net \cite{chen2019collaborative} (1 GPU) & 0.696 & 15M & $2\,170$ & $3.0550 \pm 0.0240$ \\ \hline
\end{tabular}
}%
\caption{Comparison results with other high-resolution segmenation network architectures FCtL \mbox{\cite{li2021contexts}} and GL-Net \mbox{\cite{chen2019collaborative}}. Note that the DDU-NetResNet on $1\times1$ subdomain is similar to a baseline ResNet-UNet and the implementation for other network architectures is only suitable for 1 GPU by design, in contrast to the DDU-Net. The depth of the used encoder-decoder networks is $D = 4$ blocks. $F = 256$ feature maps were communicated through the (coarse) communication network.}
\label{tab:comparison}
\end{table*}

\section{Conclusion}
\label{sec:conclusion}
This paper develops a new domain decomposition-based U-Net (DDU-Net) architecture for semantic segmentation tasks. Our results show that by including communication between subimages, the DDU-Net can handle high-resolution image segmentation efficiently without sacrificing accuracy or memory efficiency. Our approach improves segmentation accuracy by leveraging inter-subimage communication. Future research will focus on refining communication strategies, applying DDU-Net to more complex datasets, and further evaluating the benefits of parallelization on computing times and memory usage. Another direction is extending the parallelization strategy developed in this paper to other encoder-decoder architectures.

\section*{Acknowledgment}
\label{sec:acknowledgment}
The authors acknowledge the use of computational resources of the DelftBlue supercomputer, provided by Delft High Performance Computing Centre (https://www.tudelft.nl/dhpc) \cite{DHPC2024}.
This work performed in part at Sandia National Laboratories was supported by the U.S. Department of Energy, Office of Science, Office of Advanced Scientific Computing Research through the SEA-CROGS project in the MMICCs program.
Sandia National Laboratories is a multimission laboratory managed and operated by National Technology \& Engineering Solutions of Sandia, LLC, a wholly owned subsidiary of Honeywell International Inc., for the U.S. Department of Energy’s National Nuclear Security Administration under contract DE-NA0003525. This paper describes objective technical results and analysis. Any subjective views or opinions that might be expressed in the paper do not necessarily represent the views of the U.S. Department of Energy or the United States Government.

\appendix[Networks used for synthetic dataset]
\section*{Appendix}
\label{appendix:appendix-a}

\begin{table*}[ht]
\centering
\begin{tabular}{@{}lrccc@{}}
\toprule
 & \textbf{depth} $D$ & \textbf{channel distribution} & \textbf{bottleneck size} & \textbf{\# of parameters} \\ \midrule
\multirow{3}{*}{\textbf{subnetworks}} & 2 & 1-4-8-16-16-8-4-3 & $8 \times 8$ & 7\,487 \\
 & 3 & 1-4-8-16-32-32-16-8-4-3 & $4 \times 4$ & 30\,470 \\
 & 4 & 1-4-8-16-32-64-64-32-16-8-4-3 & $2 \times 2$ & 122\,031 \\ \midrule
 & \begin{tabular}[c]{@{}c@{}}\# \textbf{of feature maps}\\ \textbf{communicated} ($F$)\end{tabular} & \textbf{channel distribution} &  & \# \textbf{of parameters} \\ \midrule
\multirow{7}{*}{\textbf{coarse network}} & 1 & 1-1-1-1 &  & 84 \\
 & 2 & 2-2-2-2 &  & 318 \\
 & 4 & 4-4-4-4 &  & 1\,236 \\
 & 8 & 8-8-8-8 &  & 4\,872 \\
 & 16 & 16-16-16-16 &  & 19\,344 \\
 & 32 & 32-32-32-32 &  & 77\,088 \\
 & 64 & 64-64-64-64 &  & 307\,776 \\ \bottomrule
\end{tabular}%
\caption{Properties of the encoder-decoder network and coarse network used in the synthetic datset experiments.}
\label{tab:subnets-synthetic}
\end{table*}

\begin{table*}[ht]
\centering
\begin{tabular}{@{}lrccr@{}}
\toprule
\multicolumn{1}{c}{} & \textbf{module} & \textbf{channel distribution} & \textbf{output feature map size} & \textbf{\# of parameters} \\ \midrule
\multirow{6}{*}{\textbf{encoder-decoder network}} & ResNet18-block1 & 3-64 & $128\times128$ & 9\,536 \\
 & ResNet18-block2 & 64-64-64-64-64 & $64\times64$ & 147\,968 \\
 & ResNet18-block3 & 64-128-128-128-128 & $32\times32$ & 525\,568 \\
 & ResNet18-block4 & 128-256 & $16\times16$ & 2\,099\,712 \\
 & inter-conv ($3\times3$) & 256-256-256 & $16\times16$ & 1\,180\,672 \\
 & up-sampling path & 256-128-64-32-7 & $256\times256$ & 845\,415 \\ \midrule
\multicolumn{1}{c}{} & \begin{tabular}[c]{@{}c@{}}\# \textbf{of feature maps}\\ \textbf{communicated} ($F$) \end{tabular} & \textbf{channel distribution} & \textbf{kernel size} & \# \textbf{of parameters} \\ \midrule
\multirow{5}{*}{\textbf{coarse network}} & 1 & 1-1-1-1 & $5\times5$ & 0 \\
 & 4 & 4-4-4-4 & $5\times5$ & 84 \\
 & 16 & 16-16-16-16 & $5\times5$ & 19\,344 \\
 & 64 & 64-64-64-64 & $5\times5$ & 307\,776 \\
 & 256 & 256-256-256-256 & $5\times5$ & 4\,917\,504 \\ \bottomrule
\end{tabular}
\caption[Encoder-decoder network and coarse network properties for DeepGlobe land cover classification dataset]{Properties of the encoder-decoder network and coarse networks used for the DeepGlobe land cover classification segmentation Dataset.}
\label{tab:subnets-deepglobe}
\end{table*}

\begin{table}[ht]
\centering
\begin{tabular}{@{}lr@{}}
\toprule
\multicolumn{2}{c}{\textbf{training hyperparameters}} \\ \midrule
learning rate            & 0.008      \\
(max.) number of epochs         & 40         \\
early stopping patience         & 8 epochs   \\
learning rate decay patience    & 3 epochs   \\
minimum stopping epoch          & 20         \\
dropout rate                    & 0.1        \\
batch size                      & 16         \\ 
loss function                   & dice loss (\cref{eq:dice-loss}) \\
\bottomrule
\end{tabular}
\caption[Hyperparameters for synthetic data experiments]{Hyperparameters used for synthetic data experiments.}
\label{tab:hyperparams1}
\end{table}

\begin{figure}[ht]
    \centering
    \includegraphics[width=1.0\linewidth]{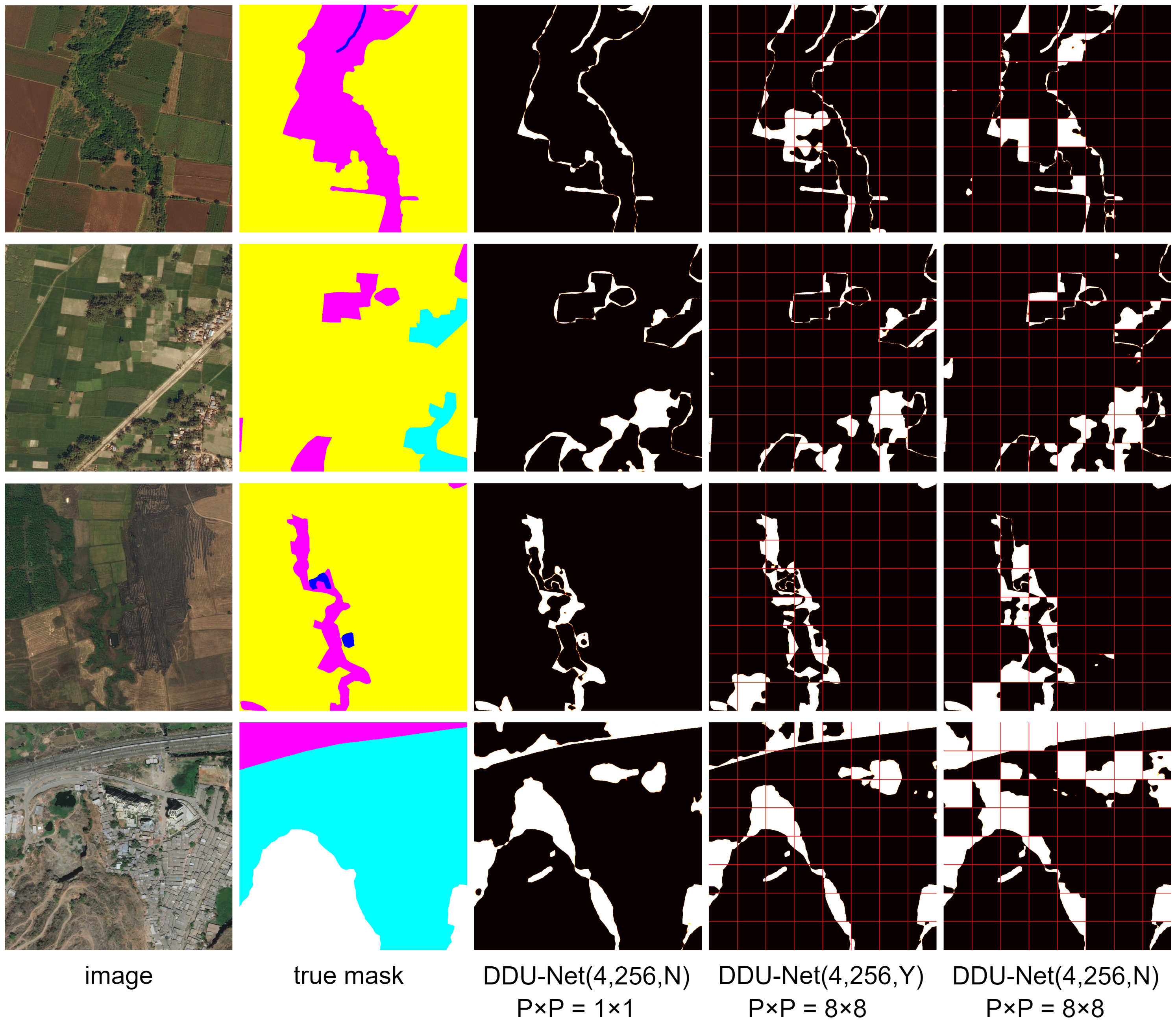}
    \caption{Visualization of prediction errors for various network configurations of the DDU-Net. Black indicates correct predictions, white indicates incorrect predictions. From left to right: original image, true mask, error by the \DDUNet{4}{256}{N} evaluated on the full image, error by the \DDUNet{4}{256}{Y} evaluated on $8 \times 8$ subimages, and error by the \DDUNet{4}{256}{N} also evaluated on $8 \times 8$ subimages. The borders of the $8 \times 8$ subimages are indicated by red lines. Note that these predictions correspond to the ones shown in~\cref{fig:qualitative-deepglobe-predictions}.}
    \label{fig:qualitative-deepglobe-predictions-differences}
\end{figure}

The properties of the encoder-decoder and coarse networks used for the synthetic dataset experiments are detailed in~\cref{tab:subnets-synthetic}. The properties are shown for different configurations in terms of encoder-decoder depth $D$. The numbers of weights for the encoder-decoder and coarse network architectures used for the DeepGlobe land cover classification dataset are detailed in~\cref{tab:subnets-deepglobe}. The encoder-decoder network consists of multiple ResNet18 blocks followed by an inter-convolutional and up-sampling path. We show the number of weights in the coarse network for various choices of its architecture, corresponding to the different experiments as shown, for instance, in~\cref{fig:evaluation-results-deepglobe}. \Cref{tab:hyperparams1} summarizes the hyperparameters employed during training for synthetic data experiments.~\Cref{fig:qualitative-deepglobe-predictions-differences} visualizes the difference between predicted and true masks for different configurations of the DDU-Net.

\bibliographystyle{IEEEtran}
\bibliography{0_additional_files/ref}

\begin{IEEEbiography}[{\includegraphics[width=1in,height=1.25in,clip,keepaspectratio]{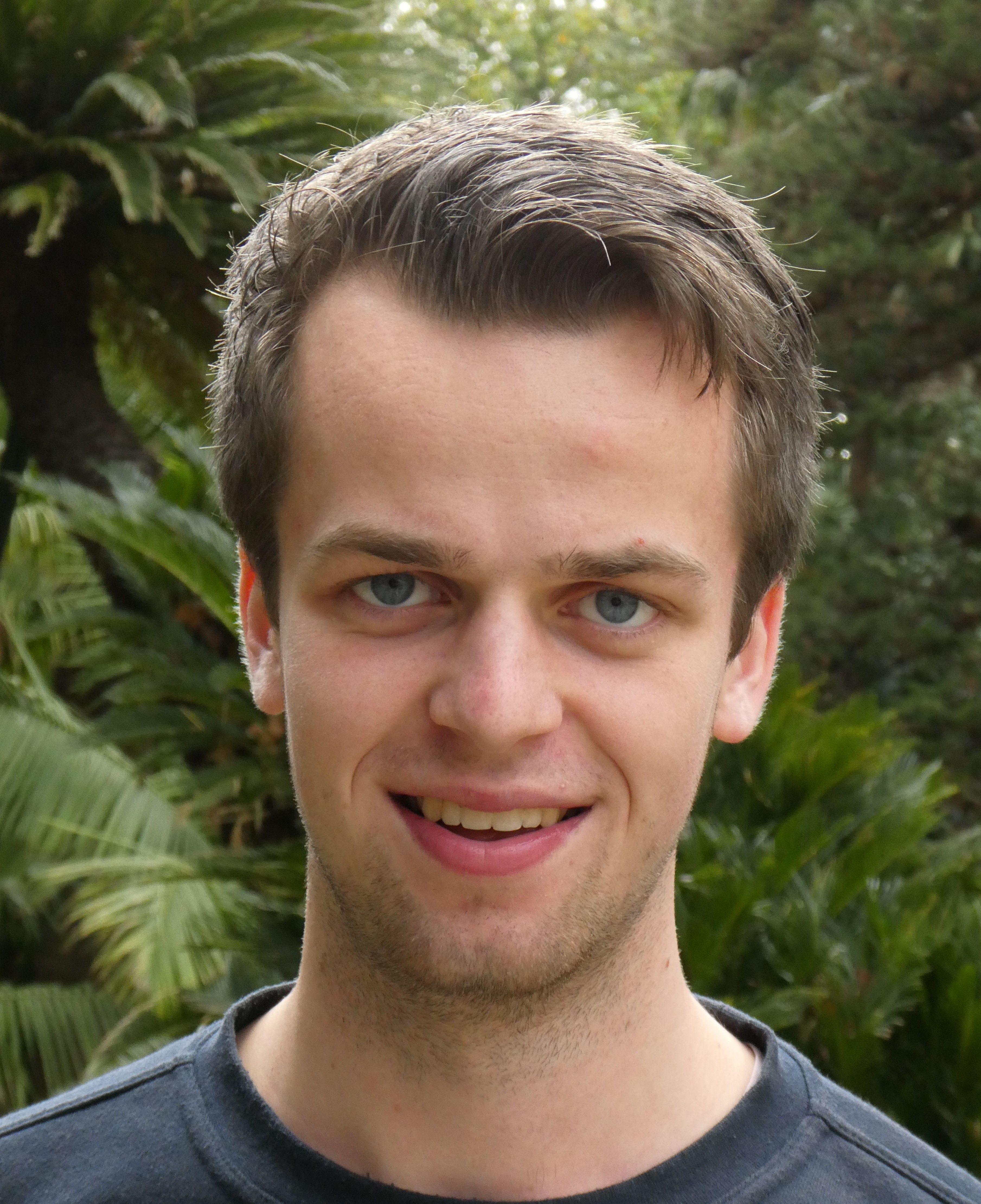}}]{Corné Verburg} received the B.S. degrees in Applied Physics and Applied Mathematics from TU Delft, the Netherlands, in 2022, and the M.S. degree in Applied Mathematics from TU Delft in 2024. He is currently pursuing the Ph.D. degree at TU Delft, focusing on biology-informed data-supported dynamic modeling and system identification for complex biological systems.

For his master's thesis, he worked on high-resolution image segmentation using convolutional neural networks (CNNs), integrating domain decomposition strategies, which resulted in the current paper. His research interests include scientific machine learning (SciML) techniques and the dynamical modeling of complex systems, particularly those that are partially unknown, using a combination of statistical and machine learning-driven methods.
\end{IEEEbiography}

\begin{IEEEbiography}[{\includegraphics[width=1in,height=1.25in,clip,keepaspectratio]{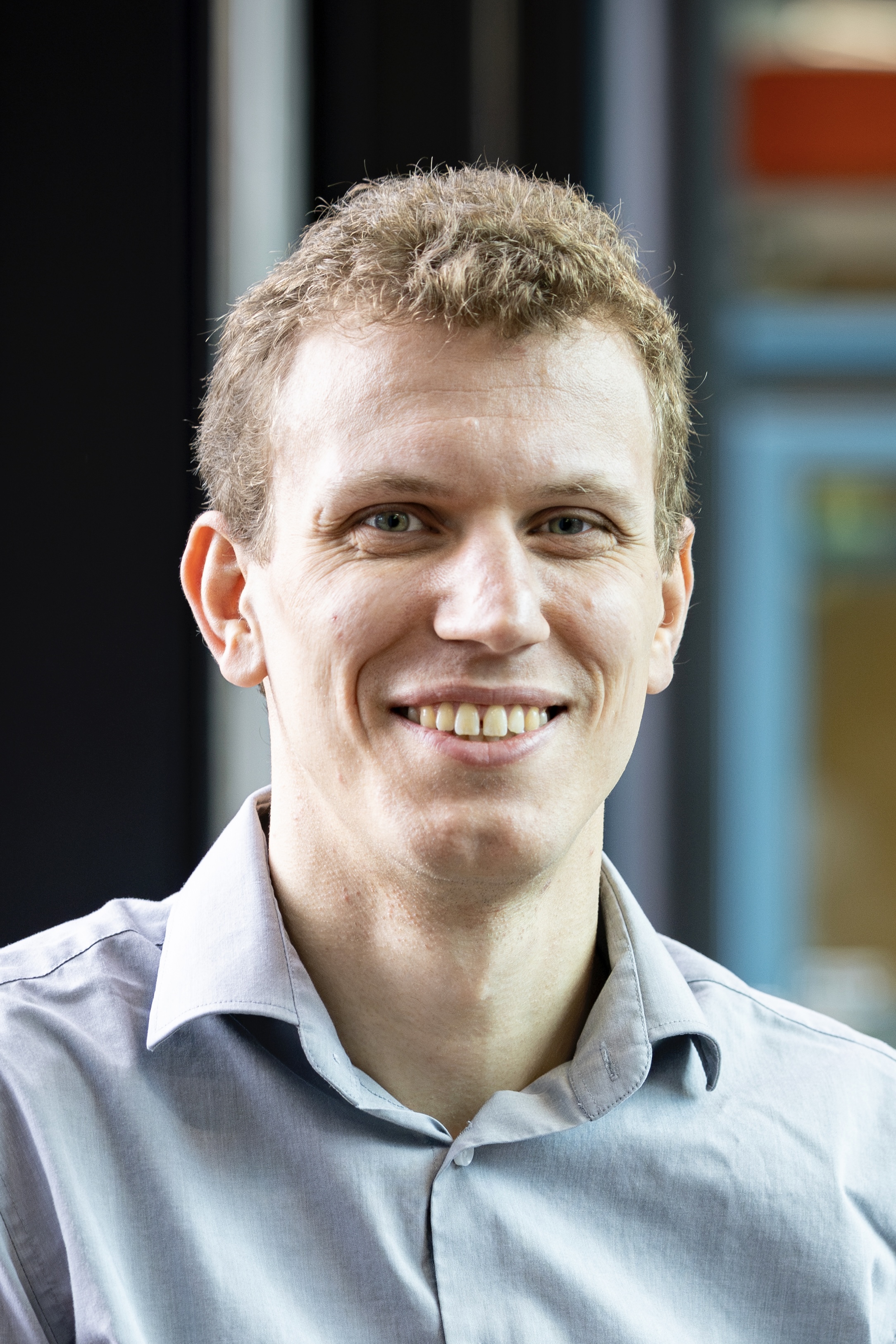}}]{Alexander Heinlein} received his diploma in mathematics from the University of Duisburg-Essen in 2011, and his doctorate degree in Mathematics from the University of Cologne in 2016. After his doctorate, he was a postdoc at the University of Cologne until 2021. From 2018 to 2021, he was the managing coordinator of the Center for Data and Simulation Science (CDS) at the University of Cologne. Before becoming an Assistant Professor for Numerical Analysis with TU Delft in 2021, he was the acting full professor for Numerical Mathematics for High Performance Computing at the University of Stuttgart. His research interests include numerical methods for partial differential equations, scientific and high-performance computing, and scientific machine learning. He focuses on domain decomposition and multiscale methods, model order reduction techniques, and their combination with machine learning. He also works on high-performance implementations on current computer architectures (CPUs, GPUs) and the application to real-world problems.
	
	Dr. Heinlein is a member of the American Mathematical Society (AMS), Dutch-Flemish Scientific Computing Society (SCS), European Mathematical Society (EMS), German Association for Computational Mechanics (GACM), Association of Applied Mathematics and Mechanics (GAMM), and Society for Industrial and Applied Mathematics (SIAM).
\end{IEEEbiography}

\begin{IEEEbiography}[{\includegraphics[width=1in,height=1.25in,clip,keepaspectratio]{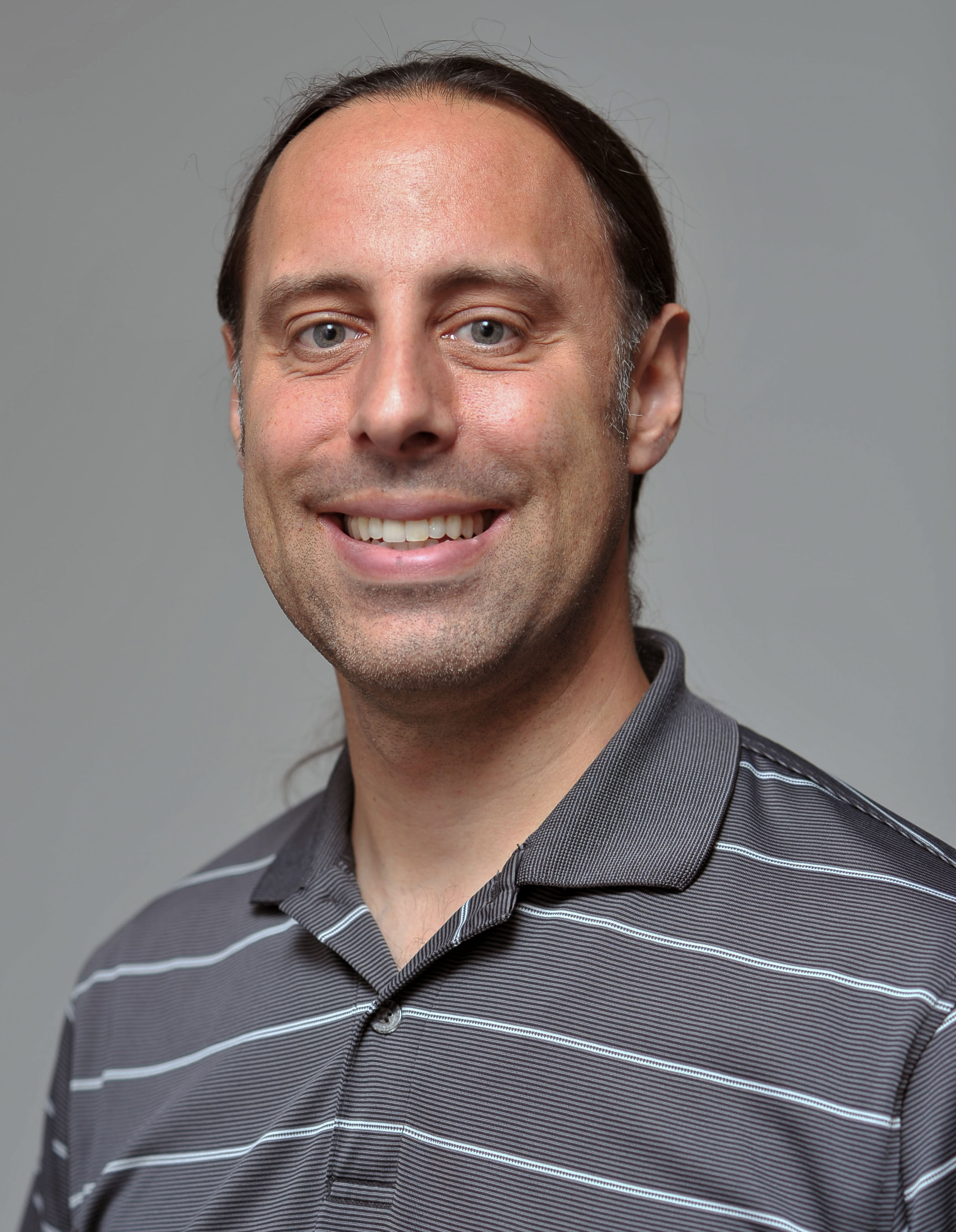}}]{Eric C. Cyr} received his B.S. in Computer Science from Clemson University in 2002, and his Ph.D. in Computer Science from University of Illinois at Urbana-Champaign in 2008. Following graduation he was a postdoctoral researcher at Sandia National Laboratories, where he has continued as a staff member since May of 2010. In 2018, he received the U.S. Department of Energy Early Career Award to develop layer-parallel methods for training deep neural networks. A subset of his research interests include scientific machine learning, parallel training algorithms, preconditioning large-scale systems, multigrid methods, and discretizations used computational plasma physics. 

Dr. Cyr is a member of the Society of Industrial and Applied Mathematics (SIAM), where he serves as associated editor for the Journal on Scientific Computing (SISC). Additionally, he has been a member of the program committee for the Copper Mountain Conference on Multigrid Methods since 2023.
\end{IEEEbiography}

\EOD

\end{document}